\documentclass[10pt,twocolumn,letterpaper]{article}

\usepackage{cvpr}
\usepackage{times}
\usepackage{epsfig}
\usepackage{graphicx}
\usepackage{amsmath}
\usepackage{amssymb}
\usepackage{algorithm}
\usepackage{algorithmicx}
\usepackage{algpseudocode}
\usepackage{threeparttable}
\usepackage{subfig}
\usepackage[authoryear]{natbib}
\usepackage{tabularx}
\usepackage{booktabs}
\usepackage[T1]{fontenc}

\usepackage[breaklinks=true,bookmarks=false]{hyperref}

\cvprfinalcopy 

\def\cvprPaperID{****} 

\algnewcommand\algorithmicforeach{\textbf{for each}}
\algdef{S}[FOR]{ForEach}[1]{\algorithmicforeach\ #1\ \algorithmicdo}

\begin{document}

\title{Hard hat wearing detection based on head keypoint localization}

\author{Bartosz Wójcik\textsuperscript{1,2}, Mateusz Żarski\textsuperscript{1,2}, Kamil Książek\textsuperscript{2,3}, Jarosław A. Miszczak\textsuperscript{2}, Mirosław J. Skibniewski\textsuperscript{4,2}\\
\small \textsuperscript{1} Faculty of Civil Engineering, Silesian University of Technology\\
\small \textsuperscript{2} Institute of Theoretical and Applied Informatics, Polish Academy of Sciences\\
\small \textsuperscript{3} Faculty of Automatic Control, Electronics and Computer Science, Silesian University of Technology\\
\small \textsuperscript{4} Department of Civil \& Environmental Engineering, University of Maryland, College Park\\
Correspondence: {\tt\small bartosz.wojcik@polsl.pl}
}

\maketitle

\begin{abstract}
    In recent years, a lot of attention is paid to deep learning methods in the context of vision-based construction site safety systems.
    However, still there is no reliable way to establish the relationship between supervised construction workers and their basic personal protective equipment, like hard hats.
    To address this problem, a novel deep learning method combining object detection, keypoint localization, and simple rule-based reasoning is proposed in this article. 
    In tests, this solution surpassed the previous methods based on the relative bounding box position of different instances, as well as direct detection of hard hat wearers and non-wearers. Achieving MS COCO style overall AP of 67.5\% compared to 66.4\% and 66.3\% achieved by the above-mentioned approaches, with class specific AP for hard hat non-wearers of 64.1\% compared to 63.0\% and 60.3\%.
    The results show that the conjunction of deep learning methods with humanly-interpretable rule-based algorithm is better suited for detection of hard hat non-wearers.\\
    Code is available at \href{https://github.com/barwojcik/hard_hats}{barwojcik/hard\_hats}.
\end{abstract}

\section{Introduction}\label{sec:introduction}

Construction is one of the most dangerous industries, together with manufacturing it leads in number of non-fatal and fatal accidents~\citep{Eurostat2020,BLS2020}. Among accidents occurring on a construction site, particularly dangerous are those in which the head is injured. While in non-fatal ones the share of head injuries is only about 7\%, in fatal they account for over 30\% of all occurrences~\citep{Eurostat2020}. This makes them a significant problem that has a key impact on the safety of construction workers.

The most common head injury occurring is TBI - traumatic brain injury \citep{Colantonio2009}. The injury itself can be fatal~\citep{Konda2016,Taylor2017} and occurs when the rapid acceleration or deceleration of the head causes the brain to move and collide with the skull. It has been identified that the most common causes of TBI's on construction sites are falls and being struck by or against an object \citep{Colantonio2009,Colantonio2010,Salem2013}.

Traditionally, workers are directly supervised on construction sites by a foreman and additional safety inspectors that are tasked with enforcing safety rules. However, continuous supervision is impossible simply due to sheer disproportion in numbers. A study performed by \cite{Bhanu2019} shows that 80\% of workers believes that work-related TBIs could be prevented, 50\% of them points out that they do not received safety training and over half of them are not supervised during work. Moreover, only one third of supervised workers thinks that supervision was adequate. This aligns well with earlier study of \cite{Liu2011}, as they pointed out that short-term employees in the construction industry are more likely to suffer TBI than short-term employees in other industries. The recurrence of TBIs are also shocking. According to \cite{Bhanu2019}, almost 40\% of questioned workers who experienced them, admits that this was not the first time they had such an injury.

\cite{Konda2016} suggested that the overall decline in TBI-related fatalities is due, among other factors, to better safety measures. They also highlighted the importance of wearing a hard hat, and the need for further improvement in head protection equipment. The latter was also stated in the recent study of \cite{BROLIN2021}. Additionally, they argued that better hard hats could reduce the effects of fall accidents.

Recognition of head injuries as a significant factor influencing the safety of the construction site has led to the legal regulation of the approach to Personal Protective Equipment (PPE) around the world~\citep{EU-OSHA1989,OSHA2004}. These regulations oblige the employer to provide personal protection measures for employees.

To ensure the appropriate usage of PPE, various methods based on wearable sensors as well as vision monitoring. \cite{Kelm2013} described a method that uses wearable RFID tags for detecting proper hard hat placement. Vision based methods use the on-site CCTV systems~\citep{Park2015} or UAVs~\citep{WaiKim2020} for obtaining the image data from the construction site, and pair it with shallow~\citep{Memarzadeh2012} or deep learning algorithms~\citep{Fang2018_2} for real-time hard hats detection.

The vision-based methods appear to be straightforward. However, this problem is more complex than it might seem. Simply finding hard hats and workers in the image is not enough.  The relationship between these instances have to be established. In fact, the real problem is not to find people that correctly wear PPE, but to find people that do not comply with the safety rules by not using it.

In this article, a novel approach to hard hat wearing detection is proposed. It couples object detection with keypoint localization, and rule-based reasoning. However, rather than using known models for human pose estimation, the model was trained to locate the person head, while finding instances of people and hard hats simultaneously. This unique problem formulation provides a way to determine the correct relationship between workers and their head protection. At the same time, it achieves this with a simple human--interpretable rules. Additionally, it overcomes drawbacks of currently used approaches as direct hard hat wearing detection suffers from high inter-class similarity, whereas solutions based on bounding box relative position lack information to reliably establish worker--hard hat relationship. This results in better performance, especially regarding detection of hard hat non-wearers. The latter is critical from a construction safety point of view, as detecting people that do not wear hard hats is the real task. We believe that this kind of work is crucial to development of reliable construction sites safety system based on deep learning.

The rest of this paper is organized as follows. In Section~\ref{sec:related} literature review regarding deep learning in the context of construction safety and hard hat wearing detection is presented, Section~\ref{sec:methods} describes the research problem, proposed solution, its implementation and training. Experiments are presented in Section~\ref{sec:experiments} and in Section~\ref{sec:comparision} comparisons with other methods have been made. Finally, all the result are discussed in Section~\ref{sec:discussion} and conclusions drawn from that with suggestions for future work are presented in Section~\ref{sec:conclusion}.

\section{Related works}\label{sec:related}

Machine learning has found applications in many fields related to Civil Engineering. It is no different in the case of construction safety, the latest deep learning applications touches a variety of its aspects \citep{Fang2020}.

\cite{Fang2018_3} proposed a deep learning based framework to detect work performed by unauthorized workers. The framework, composed out of three modules: key video clips extraction, trade recognition and worker competency judgment, is able to extract and identify activities performed on the construction site, identify workers and check in the predefined database whether they are authorized to carry out this work. \cite{Fang2018} developed a method for safety harness wearing detection. They paired Faster R-CNN \citep{Ren2017} with custom-developed CNN to detect workers and verify if safety harness is worn. \cite{Fang2019} used Mask R-CNN \citep{He2017} combined with developed Overlapping Detection Module (ODM) to recognize workers traversing structural supports to prevent falls. The presented ODM is able to determine the relationship between workers and structural supports based on mask relative positions. \cite{Zhao2019} presented an approach for safety officer trajectory tracking on the construction site. In this work, authors use YOLOv3 \citep{Redmon&Farhadi2018} for safety officer detection and Kalman filter with Hungarian matching algorithm for tracking. \cite{Wei2019} presented an approach utilizing a Spatial and Temporal Attention Pooling Network that enables worker identification. \cite{Tang2020} used human-object interaction recognition to claim whether workers wear the correct PPE during tool usage. \cite{LuoH2020} developed a real-time system capable of detecting if workers enter hazardous areas. \cite{KHAN2021} developed Mask R-CNN based object correlation detection for mobile scaffolding safety checks.

\begin{figure*}[!htbp]
	\centering
		\includegraphics[width=.75\textwidth]{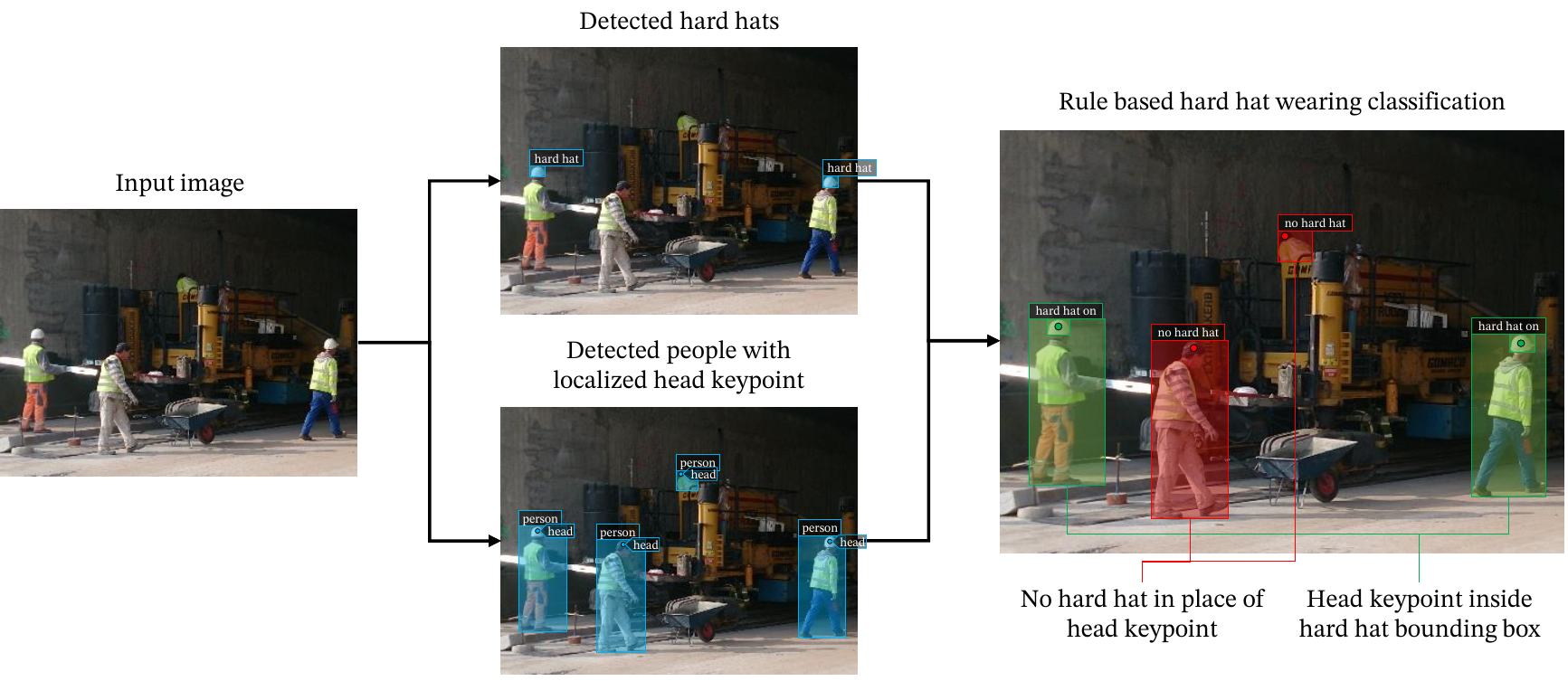}
	\caption{Workflow of the proposed solution.}
	\label{fig:proposed_solution}
\end{figure*}

\subsubsection*{Hard hat wearing detection}

Detection of hard hat wearers and non-wearers also has been addressed in recent years. \cite{Fang2018_2} used Faster R-CNN framework for this task. They also analyzed the impact of different visual conditions on the detection performance, that are specific to construction sites. \cite{Mneymneh2019} presented a multi-staged method composed of histogram of the orientated gradients and colors. In first stage workers are detected in video feed, then head protection presence in upper body part is established by object detector coupled with color based classifier. \cite{Wu2019} described a model based on Single Shot Detector framework \citep{Liu2016}, they also provided a benchmark dataset containing 3174 images. \cite{Nath2020} compared three different approaches to detecting PPE based on YOLOv3, including the following: detecting PPE and people to then establish workers -- PPE relationship based on bounding box relative position, detecting PPE wearers and non-wearers directly and finally detecting only people to determine if they are wearing PPE with different model. \cite{Lu2020} focused on real-time processing with MobilNet \citep{Howard2017} architecture. \cite{Zhou2021} tested YOLOv5 for this application.

From the collected papers, some approaches stand out from the ones referenced above. Authors of these papers tried different ways to establish the hard hat -- worker relationship. \cite{Chen2020} proposed a method based on human pose estimation and hard hat -- neck distance threshold. A euclidean distance threshold was also used by \cite{Guo2020}, this time computed between hard hat and head. The work of \cite{Shen2021} is even more interesting, as opposed to previous approaches they coupled Dual Shot Face Detector \citep{Li2019} with bounding box regression to detect if hard hat is present above workers' face. However, these methods suffers from similar drawback as both works only if certain conditions are meet. In the case of \citep{Chen2020} hips and shoulders have to be detected to compute distance form hard hat to neck and corresponding threshold. \cite{Guo2020} approach is even less flexible as the distance threshold is fixed, so it is scale dependent. On the other hand, \cite{Shen2021} solution works only when a person face is at least partially visible.

\section{Method}\label{sec:methods}
\subsection{Problem statement}

Most of the solutions used to detect the wearing of the hard hats presented in Section~\ref{sec:related} fall into one of the two general categories:
\begin{itemize}
    \item detection of people or people and hard hats, wearing a head protection is determined in different steps according to rules or another models,
    \item detecting hard hat wearers and non-wearers as separate classes.
\end{itemize}

Both categories suffer their own problems. The main problem of the first one is to establish the correct relationship between the person and the hard hat. Reasoning based on the bounding box relative position seems to be too simple to capture this, and the solutions based on distance threshold \citep{Chen2020,Guo2020} fail to set it properly for all cases.

The second category suffers from a major inter-class similarity problem. Both classes, the person wearing a hard hat and the person not wearing it, are in fact a subclass of a person class. This problem is well known in subcategory classification \citep{Cai2017,Lou2019,Han2019} as it is harder to develop a model that can correctly distinguish fine details between subcategories, in this case the presence of a hard hat. It is for this reason, the best-performing models in this group actually look for the human head instead of the whole person \citep{Fang2018_3,Wu2019,Lu2020,Zhou2021}, making them less suitable for direct transfer learning from a well-trained person detection models.

Additionally, in both cases researchers tend to disregard situations where a person is partially detected or head is obstructed, and it is not possible to tell if a person is wearing head protection. In the majority, that person is incorrectly classified as a worker without hard hat. A good example could be Pictor-v3 dataset provided by \cite{Nath2020}, in which even a person visible from waist down is annotated as hard hat non-wearer when in reality it cannot be determined. This is a serious oversight and makes it impossible to apply such a solution in practice, as the real problem is to reliably find people who are not following the safety rules.

\begin{figure*}[!htbp]
	\centering
		\includegraphics[width=.75\textwidth]{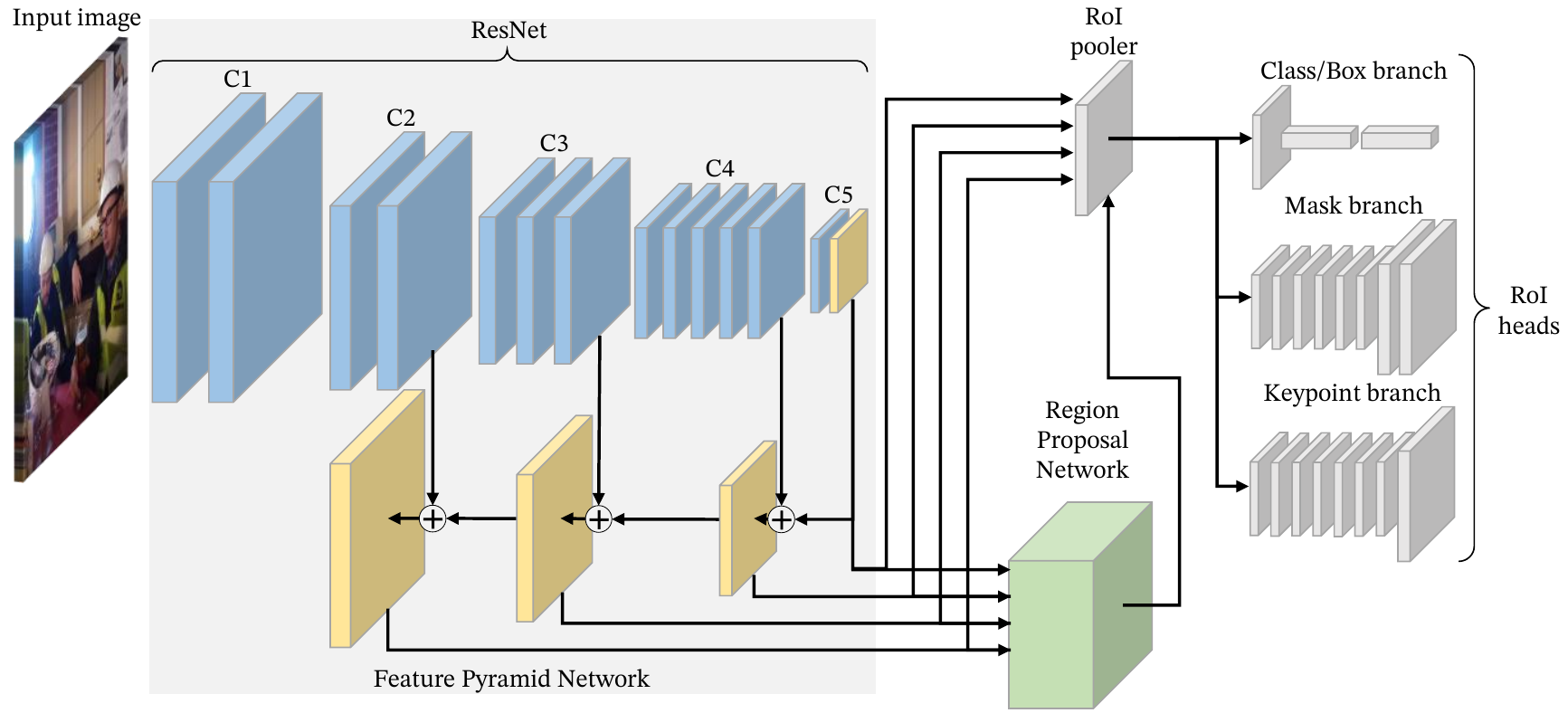}
	\caption{Generalized Region-based Convolutional Neural Network with ResNet based FPN backbone.}
	\label{fig:generalizes_rcnn}
\end{figure*}

\subsection{Keypoint based hard hat wearing detection}

To address challenges of solutions based on hard hat and person detection and incorrect partial object classification, we are introducing a new approach to detecting hard hat wearers based on head keypoint localization (Fig.~\ref{fig:proposed_solution}). In the context of deep learning, keypoints are understood as points of interest in the image. Their strongest advantage is that they are invariant for transformations, so they will not be affected by scaling.

The most common keypoint application is human pose estimation, where they represent human joints. However, instead of using an existing human pose estimation model like \cite{Chen2020} we define only one keypoint representing the localization of the human head. This model formulation enables us to correctly establish the relationship between hard hat and hard hat wearer, with a simple rule based algorithm presented in Algorithm~\ref{alg:wearing_detection}.

\begin{algorithm}[!t]
    \caption{Keypoint based hard hat wearing detection algorithm.}
        \hspace*{\algorithmicindent} \textbf{input:} $inst$ - list containing person ($p$) and hard hat\\ \hspace*{\algorithmicindent}  ($hh$) instances\\
        \hspace*{\algorithmicindent} \textbf{output:} $newInst$ - list containing person ($p$), hard hat\\
        \hspace*{\algorithmicindent} wearer ($hhw$) and hard hat non-wearer ($hhnw$)\\
        \hspace*{\algorithmicindent} instances
        \begin{algorithmic}[1]
            \ForEach{$p\ \in inst$}
                \State{$p.copyTo(newInst)$}
                \If{$p.hasHeadKp$}
                    \State{$p \to hhnw$}
                    \ForEach{$hh \in inst$}
                        \If{$p.headKp \in hh.bBox$}
                            \State{$p \to hhw$}
                            \State{\textbf{break loop}}
                        \EndIf
                    \EndFor
                \EndIf
            \EndFor\\
            \Return{$newInst$}
        \end{algorithmic}
    \label{alg:wearing_detection}
\end{algorithm}

\begin{figure}[!htbp]
	\centering
		\includegraphics[width=.75\columnwidth]{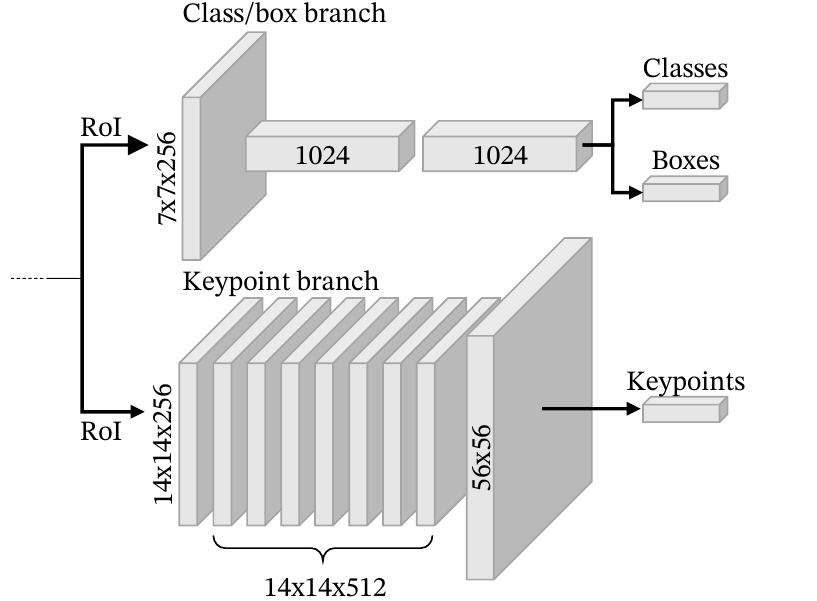}
	\caption{Keypoint R-CNN head composed of standard Faster R-CNN with FPN~\citep{Lin2017} classification/box regression branch and keypoint conv-deconv upsample branch~\citep{He2017}.}
	\label{fig:keypoint_rcnn_heads}
\end{figure}

\subsection{Implementation}

The proposed solution was implemented based on Generalized Region-based Convolutional Neural Network (Fig.~\ref{fig:generalizes_rcnn}) framework described for the first time in Mask R-CNN paper by \cite{He2017}. This natural and flexible extension to Faster R-CNN~\citep{Ren2017} enables the creation of models capable of performing a variety of tasks simultaneously. In this case, it is object detection and keypoint localization (making it Keypoint R-CNN).

\begin{table*}[!htbp]
    \caption{Breakdown of object instances in training and testing part of dataset.}
    \centering
        \begin{tabular*}{470pt}{@{\extracolsep\fill}lcccc@{\extracolsep\fill}}\toprule
            \textbf{Instances} & \textbf{all}  & \textbf{small}  & \textbf{medium}  & \textbf{large}\\
            \midrule
            \textit{Train:}\\
            hard hat & 17,741 & 11,340 & 5,922 & 479\\
            person & 23,882 & 2,805 & 9,729 & 11,348\\
            - w\slash head keypoint & 22,983 & 2,602 & 9,232 & 11,149\\
            - w\slash head keypoint wearing a hard hat & 16,700 & 1,715 & 6,459 & 8,526\\
            \midrule
            \textit{Test:}\\
            hard hat & 5746 & 3727 & 1841 & 178\\
            person & 7992 & 1077 & 3200 & 3715\\
            - w\slash head keypoint & 7775 & 1036 & 3065 & 3674\\
            - w\slash head keypoint wearing a hard hat & 5353 & 509 & 2071 & 2773\\
            \bottomrule
        \end{tabular*}
        \begin{tablenotes}
            \item Where:
            \item all - sum of all available instances in dataset
            \item small - instances with bounding box area smaller than 1024 px
            \item medium - instances with bounding box area between 1024 and 9216 px
            \item large - instances with bounding box area greater than 9216 px
        \end{tablenotes}
    \label{tab:dataset}
\end{table*}

\subsubsection*{Architectures}

Three models were implemented, each with a different backbone network featuring Feature Pyramid Network~\citep{Lin2017}. Two of those backbones were build using ResNet~\citep{He2016} architecture with layers depth of 50 and 101 (denoted as R50 and R101) and one with ResNeXt~\citep{Xie2017} architecture with layers depth of 101, block cardinality of 32 and depth of 8 (denoted as X101).

A network head was built combining standard Faster R-CNN with FPN classification and box regression branch proposed in \citep{Lin2017} with keypoint conv-deconv upscaling branch described in \citep{He2017}. Detailed architecture of Keypoint R-CNN head used is shown in Fig.~\ref{fig:keypoint_rcnn_heads}.

\subsubsection*{Dataset}

\begin{figure}[!htbp]
	\centering
		\includegraphics{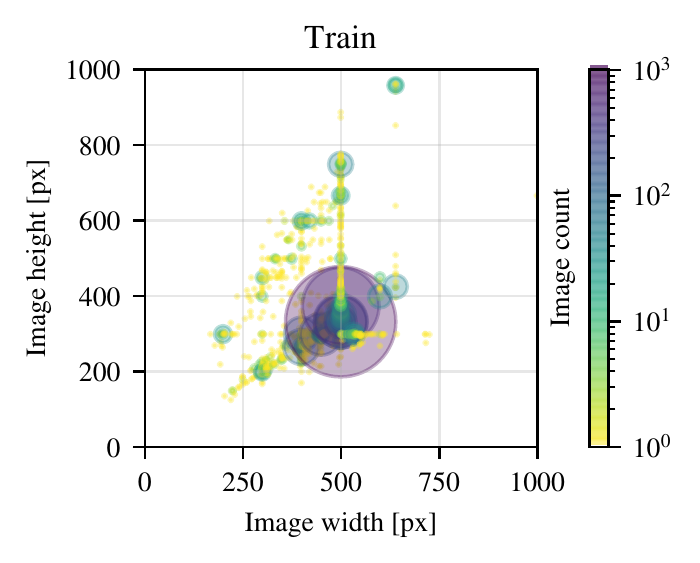}
		\includegraphics{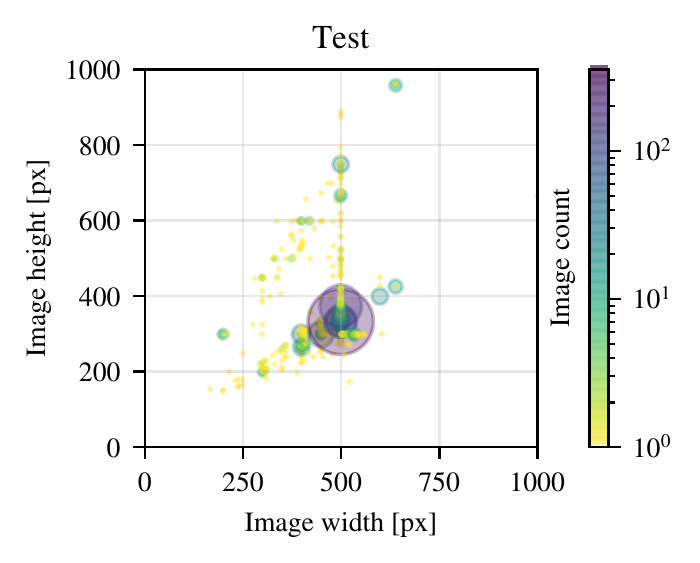}
	\caption{Image size distribution in training and testing part of dataset~\citep{DVN/7CBGOS_2019}.}
	\label{fig:dataset_sizes}
\end{figure}

Publicly available dataset~\citep{DVN/7CBGOS_2019} was used for the training and testing of our solution. This dataset contains 7035 images of different sizes, split into train (5269 images) and test (1766 images) part. The average image size is 358 $\times$ 476 px for train and 360 $\times$ 480 px for test part, with images of size 332 $\times$ 499 px being the largest group in both parts. A full breakdown of image size distribution in both parts of the dataset is presented in Fig~\ref{fig:dataset_sizes}.

As available annotations were incompatible with our solution it has been labelled for hard hat and person with head keypoint detection, which resulted in over 55 thousands object instances in MS COCO format~\citep{coco2015}. A detailed breakdown for training and testing dataset, broken down by category and sub-category with accordance to the bounding box area is presented in Tab.~\ref{tab:dataset}.

Compared to original annotations, ours contained almost 4 thousands more person instances in training and over 1.3 thousands more person instances in testing part. This difference probably comes from the number of small size instances, as the dataset originally contained annotations of people heads that are smaller than the silhouette of a whole person.

\begin{figure*}[!htbp]
	\centering
		\includegraphics[width=\textwidth]{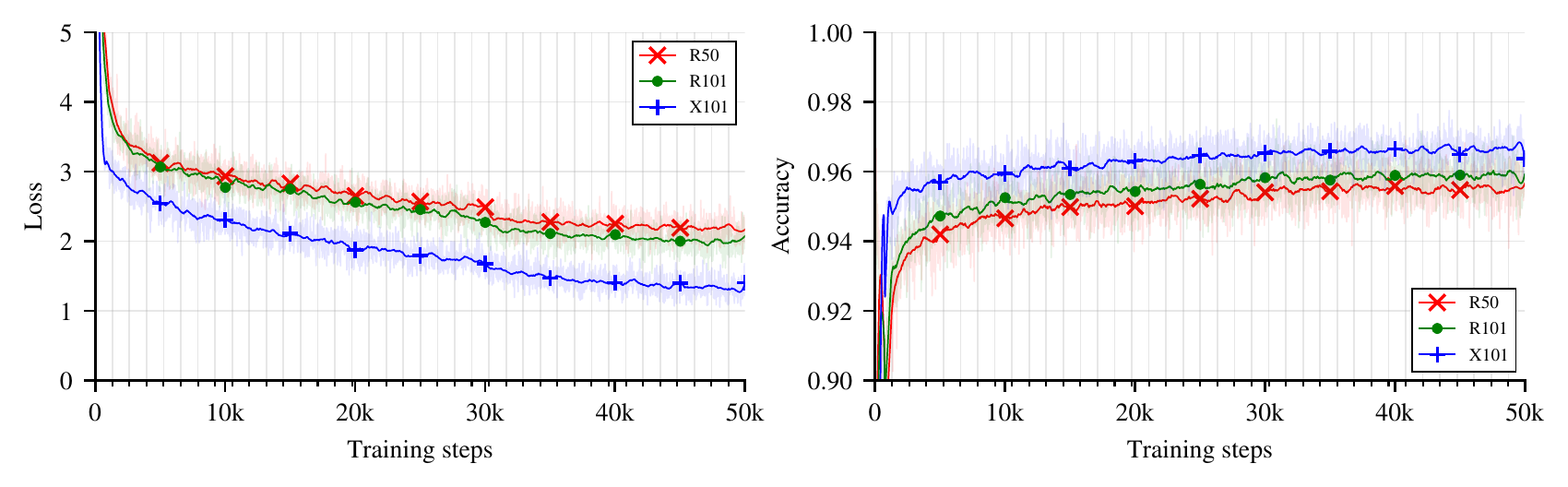}
	\caption{Value of loss function (left) and (right) accuracy on the training set throughout training for all models. Solid color lines represent smoothed values whereas transparent ones raw data. Additionally, vertical grid lines mark the end of the training epoch.}
	\label{fig:training_metrics}
\end{figure*}

\begin{figure*}[!htbp]
	\centering
		\includegraphics[width=\textwidth]{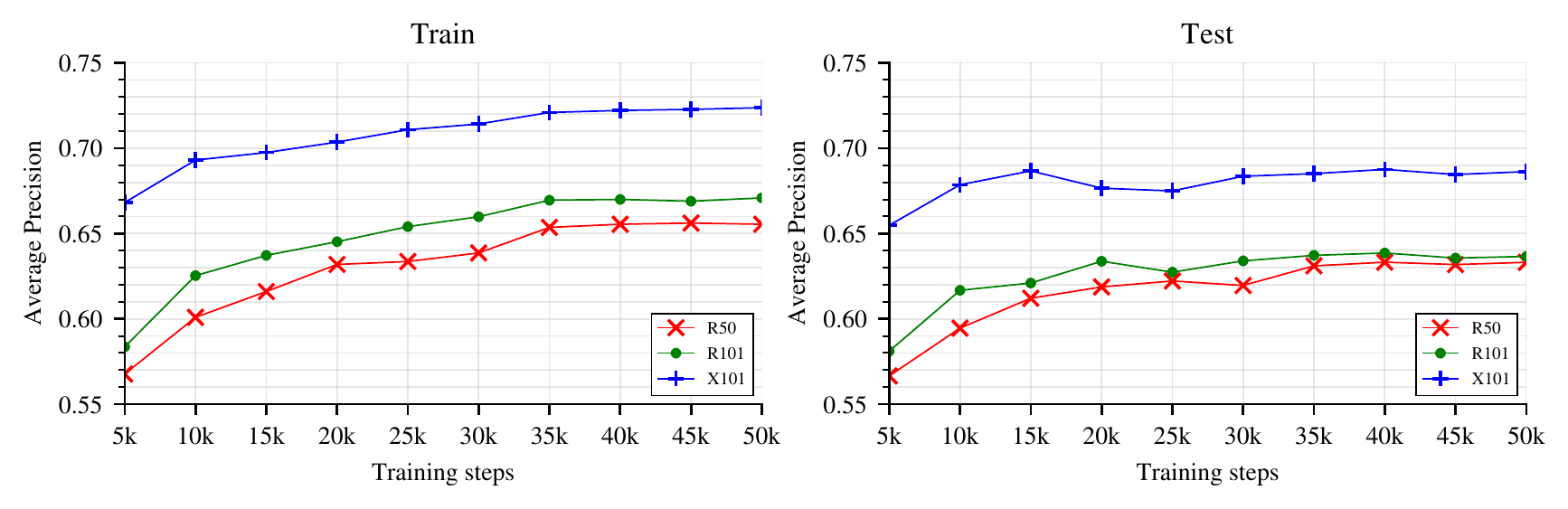}
	\caption{MS COCO style AP of each model computed every 5 thousands training steps on the train (left) and test (right) dataset.}
	\label{fig:training_eval}
\end{figure*}

\subsubsection*{Training}

Transfer learning, which is a popular technique in deep learning, was used to accelerate training. All backbones were initiated from model weights trained in human pose estimation for around 37 epochs on MS COCO 2017 dataset \citep{wu2019detectron2}, on which they achieved scores close to state-of-the-art models. Additionally, the first two layers of the backbone were frozen, as they extract general features that do not have to be retrained.

Each model was then trained on the annotated train part of \cite{Xie2017} dataset for 50 thousand steps with batch size of 4, resulting in almost 38 training epochs. Following data augmentation was used: images could be randomly flipped in horizontal, vertical or in both axes simultaneously. Furthermore, the shorter edge was randomly resized to 640, 672, 704, 736, 768 or 800 pixels. At the same time, the dimension of the longer edge could not exceed 1333 pixels. The value of loss function and classification accuracy measured for each model throughout training steps is shown in Fig.~\ref{fig:training_metrics}.

The hyperparameters were set according to \citep{wu2019detectron2}, thus the original data set divided into training and testing, was kept. Instead of monitoring loss function value on validation dataset, models were evaluated each 5 thousand training iterations on both training and testing dataset to ensure lack of overfitting in final model. MS COCO style AP (see Section~\ref{sec:experiments}) for all models computed on train and test dataset are shown in Fig.~\ref{fig:training_eval}.

\section{Experiments}\label{sec:experiments}

To evaluate trained models, a series of experiments were performed using the test part of the dataset. Additionally, due to significant size variations of dataset and already used augmentation in training, the shorter edge of the test images were resized to 800 pixels, but at the same time the dimension of the longer edge could not exceed 1333 pixels, during inference on test dataset.

\subsection{Bounding box detection}

\subsubsection*{Evaluation metrics}

Average precision (AP) and mean average precision (mAP) are the most commonly used metrics for the evaluation of object detectors. Both metrics were developed to address a need to quantify both classification and localization performance at the same time. AP averages precision (p) values across recall (r) range for a specific class, whereas mAP provides an overall metric by averaging APs for the collection of classes. Where precision
\begin{align}
    p = \frac{\mathrm{true\ positives}}{\mathrm{true\ positives} + \mathrm{false\ positives}}
\end{align}
measures percentage of correct predictions out of all predictions and recall
\begin{align}
    r = \frac{\mathrm{true\ positives}}{\mathrm{true\ positives} + \mathrm{false\ negatives}}
\end{align}
percentage of instances found.

AP has a value from 0 to 1 as both precision and recall fall in the same range and can be interpreted as area under precision-recall curve. Thus, AP can be defined by the following formula
\begin{align}
    \label{eq:ap}
    \mathrm{AP} = \int_{0}^{1} p(r)dr.
\end{align}
Therefore, mAP also ranges from 0 to 1 and can be calculated accordingly
\begin{align}
    \mathrm{mAP} = \frac{1}{n} \sum_{i=1}^{n}\mathrm{AP}_{i},
\end{align}
where AP\textsubscript{i} is AP calculated for is $i$--th element of n element collection of classes.   

The theoretical formula for AP calculation presented in Equation~\ref{eq:ap} is impractical as it requires the precision-recall relationship to be a known continuous function. Instead, the precision-recall curve is approximated by sampling precision values for defined recall thresholds and AP is calculated by numerical integration. Additionally, precision values are interpolated to reduce the influence of small variations in instance rankings. This interpolated AP (AP\textsubscript{int}) can be expressed by the following formula
\begin{align}
    \label{eq:apint}
    \mathrm{AP}_{int} = \frac{1}{m} \sum_{i=1}^{m} p_{int}(r_{thr,i}),
\end{align}
where, r\textsubscript{thr,i} is $i$--th recall threshold out of m element recall threshold collection and interpolated precision
\begin{align}
    p_{int}(r_{thr}) = \max_{\widetilde{r}_{thr} \geq r_{thr}}p(\widetilde{r}_{thr})
\end{align}
is the maximal precision value out of precision values achieved at recall thresholds equal or greater to r\textsubscript{thr}.

However, for the detection performance measurements in this article, MS COCO style AP metrics were used. These are stricter and thus provides more insight into the detector performance. Traditionally, AP and mAP metrics are computed at intersection over union (IoU) of 50\%. That means that detection is treated as positive if the ratio between common part of its bounding box and ground truth (intersection) and area encompassed by both (union) is greater than or equal to 0.5. Whereas MS COCO style AP (AP\textsubscript{COCO}) averages 101-point AP\textsubscript{int} over ten IoU thresholds, from 50\% to 95\% with the step of 5\%. This cloud be expressed by the following formula
\begin{align}
    \mathrm{AP}_{COCO} = \frac{1}{j} \sum_{i=1}^{j} \mathrm{AP}_{int,i},
\end{align}
where, AP\textsubscript{int,i} is interpolated AP defined in Equation~\ref{eq:apint} computed for $i$--th IoU threshold out of j element collection of IoU thresholds.

MS COCO style metrics also includes AP computed at IoU of 50\% and 75\%, denoted as AP\textsuperscript{50} and AP\textsuperscript{75}, as well as AP computed for objects at different scales, denoted as AP\textsubscript{S}, AP\textsubscript{M} and AP\textsubscript{L}. Moreover, creators of these metrics abandoned the distinction between AP and mAP as both are in fact a mean value but computed over different collections. Instead, the difference between these should be well stated in the context.

\begin{table*}[!htbp]
    \caption{Bounding box detection results.}
    \centering
        \begin{tabular*}{470pt}{@{\extracolsep\fill}lcccccc@{\extracolsep\fill}}\toprule
            \textbf{Model} & \textbf{AP}  & \textbf{AP\textsuperscript{50}}  & \textbf{AP\textsuperscript{75}}  & \textbf{AP\textsubscript{S}} & \textbf{AP\textsubscript{M}} & \textbf{AP\textsubscript{L}}\\
            \midrule
            \textit{Overall:}\\
            R50 & 0.633 & 0.909 & 0.727 & 0.402 & 0.699 & 0.774 \\
            R101 & 0.637 & 0.912 & 0.721 & 0.401 & 0.706 & 0.778 \\
            X101 & \textbf{0.686} & \textbf{0.940} & \textbf{0.782} & \textbf{0.466} & \textbf{0.756} & \textbf{0.820} \\
            \midrule
            \textit{Hard hat:}\\
            R50 & 0.599 & 0.900 & 0.700 & 0.511 & 0.756 & 0.771 \\
            R101 & 0.591 & 0.901 & 0.679 & 0.499 & 0.752 & 0.765 \\
            X101 & \textbf{0.626} & \textbf{0.927} & \textbf{0.731} & \textbf{0.536} & \textbf{0.776} & \textbf{0.801} \\
            \midrule
            \textit{Person:}\\
            R50 & 0.667 & 0.918 & 0.753 & 0.294 & 0.643 & 0.777 \\
            R101 & 0.682 & 0.923 & 0.763 & 0.303 & 0.661 & 0.792 \\
            X101 & \textbf{0.746} & \textbf{0.953} & \textbf{0.833} & \textbf{0.395} & \textbf{0.736} & \textbf{0.839} \\
            \bottomrule
        \end{tabular*}
        \begin{tablenotes}
            \item Where:
            \item AP - MS COCO style AP computed at different IoUs (from 50\% to 95\% with a step of 5\%)
            \item AP\textsuperscript{50}, AP\textsuperscript{75} - AP computed at IoU of 50\% and 75\%
            \item AP\textsubscript{S}, AP\textsubscript{M}, AP\textsubscript{L} - AP computed for small, medium and large objects (see Tab. \ref{tab:dataset})
        \end{tablenotes}
    \label{tab:detection}
\end{table*}

\begin{table*}[!htbp]
    \caption{Person head keypoint localization results.}
    \centering
        \begin{tabular*}{470pt}{@{\extracolsep\fill}lcccccc@{\extracolsep\fill}}\toprule
            \textbf{Model} & \textbf{AP}  & \textbf{AP\textsuperscript{50}}  & \textbf{AP\textsuperscript{75}}  & \textbf{AP\textsubscript{M}} & \textbf{AP\textsubscript{L}}\\
            \midrule
            R50& 0.704 & 0.814 & 0.736 & 0.697 & 0.854 \\
            R101 & 0.707 & 0.819 & 0.740 & 0.705 & 0.856 \\
            X101 & \textbf{0.747} & \textbf{0.838} & \textbf{0.767} & \textbf{0.748} & \textbf{0.884} \\
            \bottomrule
        \end{tabular*}
        \begin{tablenotes}
            \item Where:
            \item AP - MS COCO style mean AP computed at different OKSs (from 50\% to 95\% with a step of 5\%)
            \item AP\textsuperscript{50}, AP\textsuperscript{75} - AP computed at OKS of 50\% and 75\%
            \item AP\textsubscript{M}, AP\textsubscript{L} - AP computed for medium and large objects (see Tab. \ref{tab:dataset})
        \end{tablenotes}
    \label{tab:keypoint}
\end{table*}

\subsubsection*{Detection results}

Detection results show that models was properly trained, as all three perform well, achieving AP\textsuperscript{50} over 90\% and AP over 60\%. The model denoted as \emph{R50} performed worst with \emph{R101} slightly ahead and the \emph{X101} being clearly the best of the considered ones as it seems that deeper backbones provide better performance. The full breakdown of the results containing overall and class specific MS COCO style metrics for each model are summarized in Tab.~\ref{tab:detection}.

Examining class specific results it can be noticed that models achieve worse AP for hard hat class, as seen in Tab.~\ref{tab:detection}. Better performance in person detection is no surprise, as the training part of the dataset contains more person instances. Additionally, these models was derived from models trained only for person detection. Considering the above, this bias would be expected. However, it should be also pointed out that the difference in AP\textsuperscript{50} is not that significant and hard hats as smaller objects compared to people, achieved better scores in AP\textsubscript{S} and AP\textsubscript{M} metrics.

\subsection{Head keypoint localization}

\subsubsection*{Object Keypoint Similarity}

In MS COCO keypoints are evaluated with the similar metrics as bounding boxes. However, there is one very important distinction as opposed to bounding boxes -- IoU cannot be computed for point representations. Therefore, positive detection is determined with the use of object keypoint similarity (OKS) metric.

OKS computes the Euclidean distance between the detected keypoint and its ground truth, normalized by the scale of the bounding box. The exact formula can be expressed in the following manner
\begin{align}
    OKS = \frac{\sum_{i}\exp\{-d_i^2/2s^2k_i^2\}\delta(\vartheta_i>0)}{\sum_{i}\delta(\vartheta_i>0)},
\end{align}
where $s$ is an object scale computed from the bounding box, $d_i$ is keypoint-ground truth distance for the $i$--th keypoint, $\vartheta_i$ is visibility flat that takes positive values if $i$--th keypoint is indicated. A value $k_{i}$ is a constant specific to the $i$--th keypoint, according to the following formula
\begin{align}
    k_i = 2\sigma_i.
\end{align}
The value $\sigma_i$ is the $i$--th keypoint standard deviation computed relative to the object scale, over a set of redundantly annotated images. In our case the head keypoint $\sigma$ value was set to 0.026 according to the value provided for AI Challenge Keypoint Dataset \citep{wu2017ai} as there was not enough redundantly annotated images in our dataset to compute k constant. This value aligns well with MS COCO values form on head features.

\begin{table*}[!htbp]
    \caption{Comparison of person head keypoint localization results for person with and without hard hat.}
    \centering
        \begin{tabular*}{470pt}{@{\extracolsep\fill}lcccccc@{\extracolsep\fill}}\toprule
            \textbf{Model} & \textbf{AP}  & \textbf{AP\textsuperscript{50}}  & \textbf{AP\textsuperscript{75}}  & \textbf{AP\textsubscript{M}} & \textbf{AP\textsubscript{L}}\\
            \midrule
            \textit{Person w\slash hard hat:}\\
            R50 & 0.727 & 0.809 & 0.756 & 0.697 & 0.847 \\
            R101 & 0.732 & 0.816 & 0.763 & 0.700 & 0.852 \\
            X101 & \textbf{0.774} & \textbf{0.841} & \textbf{0.799} & \textbf{0.751} & \textbf{0.883} \\
            \midrule
            \textit{Person w\slash o hard hat:}\\
            R50 & 0.536 & 0.646 & 0.567 & 0.573 & 0.742 \\
            R101 & 0.555 & 0.671 & 0.582 & 0.590 & 0.766 \\
            X101 & \textbf{0.609} & \textbf{0.713} & \textbf{0.628} & \textbf{0.650} & \textbf{0.817} \\
            \bottomrule
        \end{tabular*}
        \begin{tablenotes}
            \item For definition of AP, AP\textsuperscript{50}, AP\textsuperscript{75}, AP\textsubscript{M} and AP\textsubscript{L} see Tab. \ref{tab:keypoint}
        \end{tablenotes}
    \label{tab:keypoint_h_hh}
\end{table*}

\begin{figure*}[!htbp]
	\centering
	    \subfloat[hard hat wearers]{
	    \includegraphics[height=2.6cm]{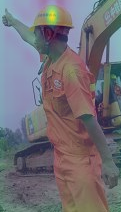}
		\includegraphics[height=2.6cm]{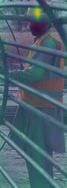}
		\includegraphics[height=2.6cm]{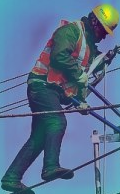}
		\includegraphics[height=2.6cm]{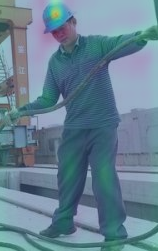}
		\includegraphics[height=2.6cm]{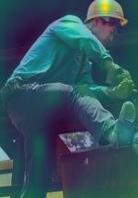}
		\includegraphics[height=2.6cm]{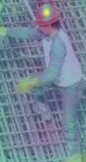}
		\includegraphics[height=2.6cm]{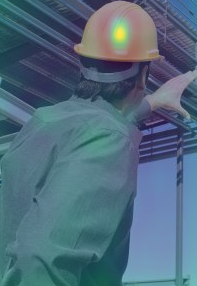}
		\includegraphics[height=2.6cm]{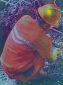}
		\includegraphics[height=2.6cm]{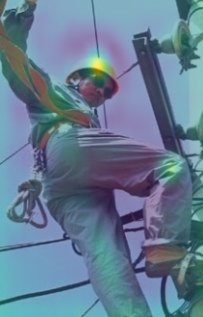}
		\includegraphics[height=2.6cm]{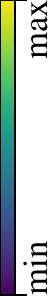}}\\
		\subfloat[hard hat non-wearers]{
		\includegraphics[height=2.6cm]{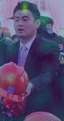}
		\includegraphics[height=2.6cm]{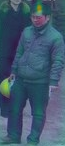}
		\includegraphics[height=2.6cm]{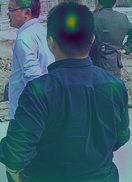}
		\includegraphics[height=2.6cm]{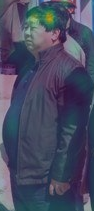}
		\includegraphics[height=2.6cm]{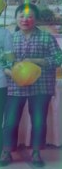}
		\includegraphics[height=2.6cm]{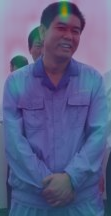}
		\includegraphics[height=2.6cm]{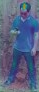}
		\includegraphics[height=2.6cm]{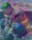}
		\includegraphics[height=2.6cm]{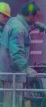}
		\includegraphics[height=2.6cm]{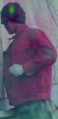}
		\includegraphics[height=2.6cm]{Figs/hh_fig7_legend.pdf}}
	\caption{Comparison of head keypoint heatmaps between hard hat wearers (a) and non-wearers (b), overlaid on object instances from test datset~\citep{DVN/7CBGOS_2019} detected by best performing model (\textit{X101}).}
	\label{fig:head_heatmaps}
\end{figure*}

\subsubsection*{Keypoint localization results}

In case of person head keypoint localization, all models also performed very good, achieving AP over 70\% and AP\textsuperscript{50} over 80\%. The results are even more impressive considering the fact that in general keypoint localization is perceived as harder task than the bounding box detection. The full breakdown of the head keypoint localization for person class is summarized in Tab.~\ref{tab:keypoint}.

\subsubsection*{Head and head with hard hat}

Apart from the above, head keypoint localization evaluation was performed on person sub-classes that represented hard hat wearers and non-wearers. This was done to check if \emph{Keypoint R-CNN} is capable of head keypoint generalization between both groups, thus could localize head keypoint whether the hard hat is worn. The full results of this evaluation are summarized in Tab.~\ref{tab:keypoint_h_hh}.

As seen in the results all models display bias towards a more prominent person sub-class representing hard hat non-wearers which is in line with instance imbalance written in Tab.~\ref{tab:dataset}.

However, head keypoint heatmaps were inspected to assure that the solution performs well. Some of these keypoint heatmaps overlaid on instances of both, hard hat wearers and non-wearers, are presented in Fig.~\ref{fig:head_heatmaps}, from these heatmaps keypoints are selected as points with the highest score. It can be seen that head keypoint is correctly localized for both groups considering different poses and scales as well as partial visibility.

\begin{table}[!htbp]
    \centering
    \caption{Detection threshold with corresponding overall F1 score for each model.}
    \label{tab:det_thr}
    \begin{tabular*}{\columnwidth}{@{\extracolsep\fill}lcc@{\extracolsep\fill}}%
        \toprule
        \textbf{Model} & \textbf{Detection threshold [\%]} & \textbf{Overall F1 score [\%]}\\
        \midrule
        R50 & 79 & 88.7\\
        R101 & 82 & 89.2\\
        X101 & 81 & 91.8\\
        \bottomrule
    \end{tabular*}
\end{table}

\subsection{Hard hat wearing}

\subsubsection*{Detection threshold moving}

\begin{table*}[!htbp]
    \caption{Results of hard hat wearing detection.}
    \centering
        \begin{tabular*}{470pt}{@{\extracolsep\fill}lcccccc@{\extracolsep\fill}}\toprule
            \textbf{Model} & \textbf{AP}  & \textbf{AP\textsuperscript{50}}  & \textbf{AP\textsuperscript{75}}  & \textbf{AP\textsubscript{S}} & \textbf{AP\textsubscript{M}} & \textbf{AP\textsubscript{L}}\\
            \midrule
            \textit{Overall:}\\
            R50 & 0.575 & 0.752 & 0.663 & 0.124 & 0.572 & 0.728 \\
            R101 & 0.595 & 0.765 & 0.681 & 0.140 & 0.587 & 0.757 \\
            X101 & \textbf{0.675} & \textbf{0.826} & \textbf{0.759} & \textbf{0.211} & \textbf{0.682} & \textbf{0.817} \\
            \midrule
            \textit{Hard hat wearer:}\\
            R50 & 0.620 & 0.823 & 0.719 & 0.167 & 0.586 & 0.723 \\
            R101 & 0.637 & 0.827 & 0.736 & 0.186 & 0.600 & 0.746 \\
            X101 & \textbf{0.710} & \textbf{0.871} & \textbf{0.805} & \textbf{0.247} & \textbf{0.693} & \textbf{0.805} \\
            \midrule
            \textit{Hard hat non-wearer:}\\
            R50 & 0.531 & 0.682 & 0.606 & 0.082 & 0.559 & 0.733 \\
            R101 & 0.553 & 0.704 & 0.626 & 0.094 & 0.574 & 0.768 \\
            X101 & \textbf{0.641} & \textbf{0.780} & \textbf{0.714} & \textbf{0.175} & \textbf{0.671} & \textbf{0.828} \\
            \bottomrule
        \end{tabular*}
        \begin{tablenotes}
            \item For definition of AP, AP\textsuperscript{50}, AP\textsuperscript{75}, AP\textsubscript{S}, AP\textsubscript{M} and AP\textsubscript{L} see Tab. \ref{tab:detection}
        \end{tablenotes}
    \label{tab:wearing_detection}
\end{table*}

The algorithm presented in Algorithm~\ref{alg:wearing_detection} does not take detection probability score into account. It makes it vulnerable to detection confidence threshold, as low scoring hard hat instances would be treated in the same manner as instances detected with nearly 100\% confidence. This means that detector used for hard hat wearing evaluation have to be properly tuned.

The confidence threshold below which objects are not treated as positive detection is called decision threshold, and the process of finding the optimal threshold is referred to as detection threshold moving. There are few strategies for this task depending on the preferences. In this case the detection threshold for each model was selected by maximization of F1 score, this was done to achieve balance between precision and recall as F1 is a harmonic mean of these metrics and can be expressed by the following formula
\begin{align}
    F1 = \frac{2(p\cdot r)}{p+r}.
\end{align}
The F1 scores were calculated for each class over the set of decision thresholds starting from 5\% to 99\% with the step of 1\%. The scores obtained this way were then averaged to get the overall F1 metric, and a threshold value with highest F1 score value was selected. The resulting decision thresholds and corresponding overall F1 scores achieved by each model are summarized in Tab.~\ref{tab:det_thr}.

\subsubsection*{Hard hat wearing results}

Once again the model denoted as \emph{X101} performed best, which was expected as classification is based on previously evaluated detection and keypoint localization. The detailed breakdown of results of hard hat wearing detection are presented in Tab.~\ref{tab:wearing_detection}.

As seen in the results, only the aforementioned model achieved AP\textsuperscript{50} over 80\% and AP over 60\% with other models closer to AP\textsuperscript{50} value of 75\% and AP below 60\%. However, to fully assess the performance of the proposed approach, it is necessary to put it into perspective by comparing it with different already presented ones.

Moreover, examining of class specific scores, it can be seen that again all models show bias towards hard hat wearer class. As stated before, it is not surprising considering the fact that in both, train and test part of the dataset, hard hat non-wearers account for about 30\% of people instances. However, troubling fact is small scale performance as in general AP\textsubscript{S} for all models did not exceed 25\% and mostly scored well below 20\%.

\section{Comparative studies}\label{sec:comparision}

Apart from experiments presented in Section~\ref{sec:experiments}, the validity of the proposed solution was also tested in comparison to other approaches characterized in Section~\ref{sec:methods}. These two comparative studies aims to:

\begin{itemize}
    \item compare our head keypoint based solution to other solution based on hard hat/person bounding box relative position,
    \item compare our solution to direct detection of hard hat wearers and non-wearers.
\end{itemize}

\begin{table*}[!htbp]
    \caption{Comparison of our keypoint based approach with proposed by \cite{Nath2020} decision tree based on the bounding box relative position.}
    \centering
        \begin{tabular*}{470pt}{@{\extracolsep\fill}lcccccc@{\extracolsep\fill}}\toprule
            \textbf{Classifier} & \textbf{AP}  & \textbf{AP\textsuperscript{50}}  & \textbf{AP\textsuperscript{75}}  & \textbf{AP\textsubscript{S}} & \textbf{AP\textsubscript{M}} & \textbf{AP\textsubscript{L}}\\
            \midrule
            \textit{Overall:}\\
            Our solution & \textbf{0.675} & \textbf{0.826} & \textbf{0.759} & 0.211 & \textbf{0.682} & \textbf{0.817} \\
            Our DT & 0.664 & 0.815 & 0.746 & \textbf{0.222} & 0.668 & 0.799 \\
            \cite{Nath2020} DT & 0.654 & 0.806 & 0.736 & \textbf{0.222} & 0.662 & 0.775 \\
            \midrule
            \textit{Hard hat wearer:}\\
            Our solution & \textbf{0.710} & \textbf{0.871} & \textbf{0.805} & 0.247 & \textbf{0.693} & \textbf{0.805} \\
            Our DT & 0.698 & 0.860 & 0.794 & 0.248 & 0.681 & 0.789 \\
            \cite{Nath2020} DT & 0.696 & 0.860 & 0.795 & \textbf{0.250} & 0.682 & 0.788 \\
            \midrule
            \textit{Hard hat non-wearer:}\\
            Our solution & \textbf{0.641} & \textbf{0.780} & \textbf{0.714} & 0.175 & \textbf{0.671} & \textbf{0.828} \\
            Our DT & 0.630 & 0.769 & 0.698 & \textbf{0.197} & 0.655 & 0.808 \\
            \cite{Nath2020} DT & 0.611 & 0.751 & 0.677 & 0.194 & 0.642 & 0.762 \\
            \bottomrule
        \end{tabular*}
        \begin{tablenotes}
            \item For definition of AP, AP\textsuperscript{50}, AP\textsuperscript{75}, AP\textsubscript{S}, AP\textsubscript{M} and AP\textsubscript{L} see Tab. \ref{tab:detection}
        \end{tablenotes}
    \label{tab:wearing_comparison}
\end{table*}

\subsection{Comparison with \cite{Nath2020} decision tree}\label{sec:comparision1}

\cite{Nath2020} described a decision tree (DT) approach for hard hat wearing classification based on the bounding box relative position. In this approach, each hard hat bounding box is normalized to the person bounding box coordinate system, then DT decides if the person is a hard hat wearer or non-wearer.

Their solution, based on YOLOv3, achieved overall AP\textsuperscript{50} value of 69.09\% for hard hat wearers/non-wearers detection, which was worse that direct detection of those classes that achieved AP\textsuperscript{50} value of 73.97\%. Regarding class specific performance, the former achieved AP\textsuperscript{50} of 74.29\% for hard hat wearer and 63.84\% for non-wearers, whereas the latter 79.81\% and 68.12\%. 

However, direct comparison is not feasible as our solution is based on different detectors that, opposed to YOLOv3, are focused on accuracy instead of real-time performance. Moreover, the dataset used by \cite{Nath2020} in their study was smaller. Therefore, to provide a fair comparison, a new DT was developed with our dataset and tested using our detector. Additionally, the aforementioned DT developed by \cite{Nath2020} was also tested this way.

The architecture of our DT was selected using grid search technique with 5-fold cross-validation on the training set. The optimized parameters were as follows:
\begin{itemize}
    \item a split criterion: Gini impurity or entropy information gain,
    \item the maximum depth of the tree $m_{d}$ to prevent overfitting: $\lbrace 2, 3, ..., 15 \rbrace$ or no limit,
    \item the minimum samples $m_{s}$ necessary to split.
\end{itemize}
Results of experiments indicate that the best set of hyperparameters is: Gini impurity criterion, $m_{d} = 10$ and $m_{s} = 14$.

The \emph{X101} model was selected as a sole detector for this comparison as it performed best among all trained ones, the results achieved are summarized in Tab.~\ref{tab:wearing_comparison}. It could be seen that \cite{Nath2020} DT achieved higher overall and class specific AP\textsuperscript{50} values on our dataset with our detector. That only underlines inability to direct comparison of methods developed in different conditions. Regarding comparison itself, our solution outperformed both \cite{Nath2020} DT and DT fitted to our dataset in almost all metrics. What is even more important, the most performance was gained in hard hat non-wearers detection, which is also true while comparing our DT to \cite{Nath2020} DT. Also, it has to be noted that DTs performed better in small scales as it achieved higher overall and class specific AP\textsubscript{S} scores.

The relative performance of both decision trees is also worth to address. As seen in Tab.~\ref{tab:wearing_comparison} overall AP is only slightly better for DT developed from scratch and regarding hard hat wearers \cite{Nath2020} DT is on par, the main difference comes from hard hat non-wearers. However, our DT is significantly more complex. It has a depth of 10 layers and is composed of 485 nodes, compared to 3 layers and 10 nodes of \cite{Nath2020} DT. This means that it lost one of the main advantage of the DTs -- human interpretability. Whereas simpler trees can be acquired, decrease in depth and node number leads to performance degradation. Given the interpretability aspect, it can be concluded that \cite{Nath2020} has already reached the limits of the decision trees.

\subsection{Comparison with direct detection of hard hat wearers}\label{sec:comparision2}

\begin{table*}[!htbp]
    \caption{Comparison of person detection results as separate category (Keypoint R-CNN) and super-category (Faster R-CNN).}
    \centering
        \begin{tabular*}{470pt}{@{\extracolsep\fill}lcccccc@{\extracolsep\fill}}\toprule
            \textbf{Model} & \textbf{AP}  & \textbf{AP\textsuperscript{50}}  & \textbf{AP\textsuperscript{75}}  & \textbf{AP\textsubscript{S}} & \textbf{AP\textsubscript{M}} & \textbf{AP\textsubscript{L}}\\
            \midrule
            Keypoint R-CNN & \textbf{0.746} & \textbf{0.953} & \textbf{0.833} & \textbf{0.395} & \textbf{0.736} & \textbf{0.839} \\
            Faster R-CNN & 0.729 & 0.940 & 0.827 & 0.369 & 0.714 & 0.830 \\
            \bottomrule
        \end{tabular*}
        \begin{tablenotes}
            \item For definition of AP, AP\textsuperscript{50}, AP\textsuperscript{75}, AP\textsubscript{S}, AP\textsubscript{M} and AP\textsubscript{L} see Tab. \ref{tab:detection}
        \end{tablenotes}
    \label{tab:person_detectrion_comparison}
\end{table*}

For the purpose of this comparison, another model based on ResNeXt backbone has been trained. Excluding the keypoint branch, it was identical to the model previously denoted as \textit{X101} and was trained starting from the same weights, with the same parameters and for the same number of training steps. However, instead of detecting hard hat and person instances, it was trained for the direct detection of hard hat wearers and non-wearers. This newly trained network was in fact the Faster R-CNN previously used for this task \citep{Fang2018_2}.

\subsubsection*{Person detection comparision}

The Tab.~\ref{tab:person_detectrion_comparison} shows a comparison of person detection evaluation between the model described above (denoted as \textit{Faster R-CNN}) and one trained in detecting hard hat and person instances (denoted as \textit{Keypoint R-CNN}, which is \textit{X101} model from Section~\ref{sec:experiments}). As seen in the table, \textit{Keypoint R-CNN} model slightly outperforms \textit{Faster R-CNN} in person detection when all instances detected by the latter one are treated as one class. The difference in performance is not that big. However, it should be noted that in general adding keypoint branch, opposed to mask branch, hinders object detection \citep{He2017}. Therefore, \emph{Faster R-CNN} should perform better than \emph{Keypoint R-CNN}.

\begin{figure}[!htbp]
	\centering
		\subfloat[Keypoint R-CNN]{\includegraphics{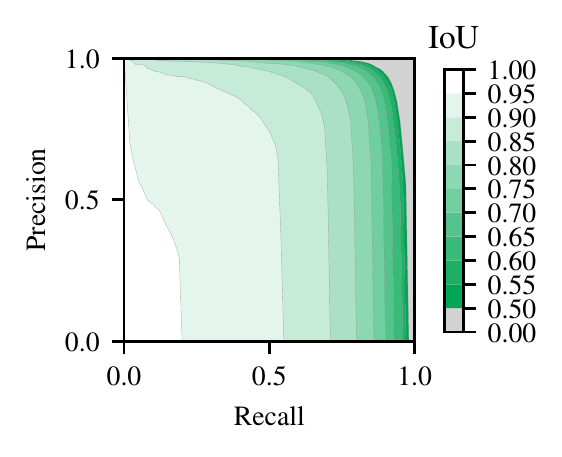}}\\
		\subfloat[Faster R-CNN]{\includegraphics{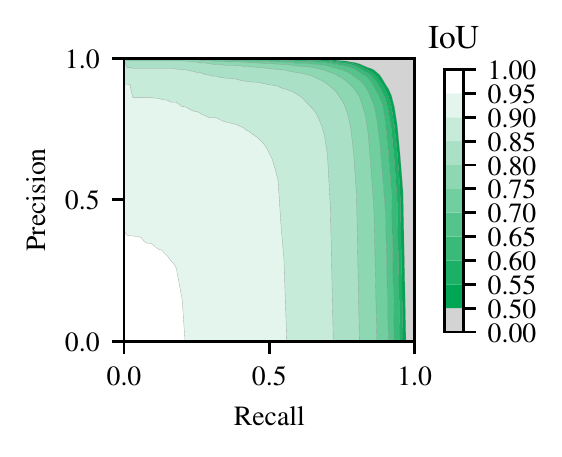}}
	\caption{Person class Precision-Recall curves for \emph{Keypoint R-CNN} (a) and \emph{Faster R-CNN} (b).}
	\label{fig:person_k_vs_f}
\end{figure}

This difference in performance should be linked to the problem mentioned earlier, high inter-class similarity. In this case the detector focuses on learning small detail, hard hat, and loses the ability to recognize more general features of a person. This is even more evident at higher IoU thresholds (Fig.~\ref{fig:person_k_vs_f}) as precision values drops faster as recall rises.

\subsubsection*{Hard hat wearing detection comparision}

To compare the final model performance in hard hat wearing detection, detection threshold moving was also performed for Faster R-CNN. In case of this model, overall F1 score achieved the highest value of 87.6\% at a decision threshold of 83\%.

The result of comparison is summarized in Tab.~\ref{tab:wearing_comparison2}. Except for AP\textsubscript{S}, \emph{Keypoint R-CNN} slightly outperformed, or in case of AP\textsuperscript{50} matched, \emph{Faster R-CNN} model. Class specific breakdown shows, that the difference in performance comes solely from hard hat non-wearer detection. It means that \emph{Keypoint R-CNN} delivers more balanced performance at the cost of hard hat wearing detection.

\begin{table*}[!htbp]
    \caption{Comparison of hard hat wearing detection.}
    \centering
        \begin{tabular*}{470pt}{@{\extracolsep\fill}lcccccc@{\extracolsep\fill}}\toprule
            \textbf{Model} & \textbf{AP}  & \textbf{AP\textsuperscript{50}}  & \textbf{AP\textsuperscript{75}}  & \textbf{AP\textsubscript{S}} & \textbf{AP\textsubscript{M}} & \textbf{AP\textsubscript{L}}\\
            \midrule
            \textit{Overall:}\\
            Keypoint R-CNN & \textbf{0.675} & \textbf{0.826} & \textbf{0.759} & 0.211 & \textbf{0.682} & \textbf{0.817} \\
            Faster R-CNN & 0.663 & \textbf{0.826} & 0.757 & \textbf{0.248} & 0.670 & 0.809 \\
            \midrule
            \textit{Hard hat wearer:}\\
            Keypoint R-CNN & 0.710 & 0.871 & 0.805 & 0.247 & 0.693 & 0.805 \\
            Faster R-CNN & \textbf{0.723} & \textbf{0.905} & \textbf{0.824} & \textbf{0.331} & \textbf{0.700} & \textbf{0.809} \\
            \midrule
            \textit{Hard hat non-wearer:}\\
            Keypoint R-CNN & \textbf{0.641} & \textbf{0.780} & \textbf{0.714} & \textbf{0.175} & \textbf{0.671} & \textbf{0.828} \\
            Faster R-CNN & 0.603 & 0.747 & 0.690 & 0.165 & 0.639 & 0.810 \\
            \bottomrule
        \end{tabular*}
        \begin{tablenotes}
            \item For definition of AP, AP\textsuperscript{50}, AP\textsuperscript{75}, AP\textsubscript{S}, AP\textsubscript{M} and AP\textsubscript{L} see Tab. \ref{tab:detection}
        \end{tablenotes}
    \label{tab:wearing_comparison2}
\end{table*}

\section{Discussion}\label{sec:discussion}

\begin{figure*}[!htbp]
	\centering
	    \subfloat[perfect detection]{\includegraphics[width=.3\textwidth]{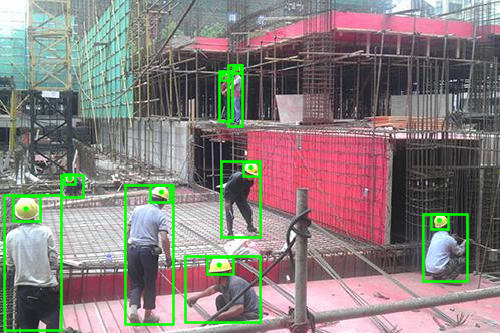}}
	    \hfill
	    \subfloat[failure to detect of one of the hard hats]{\includegraphics[width=.3\textwidth]{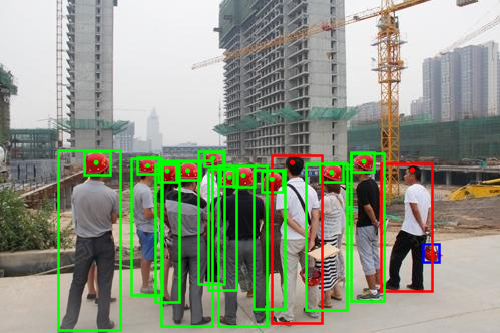}}
	    \hfill
	    \subfloat[failure to detect both person and hard hat due to occlusion]{\includegraphics[width=.3\textwidth]{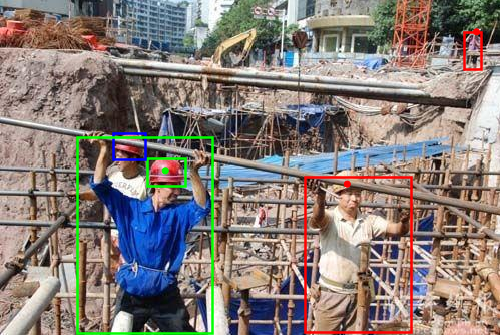}}\\
	    \subfloat[failure in keypoint localization]{\includegraphics[width=.3\textwidth]{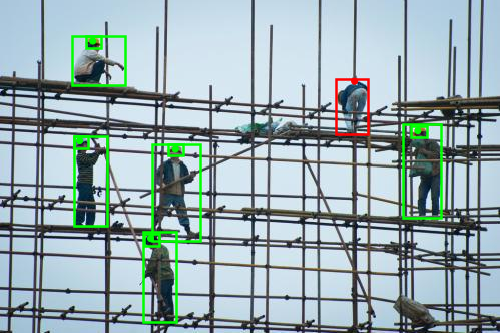}}
	    \hfill
	    \subfloat[detection failure at small scale]{\includegraphics[width=.3\textwidth]{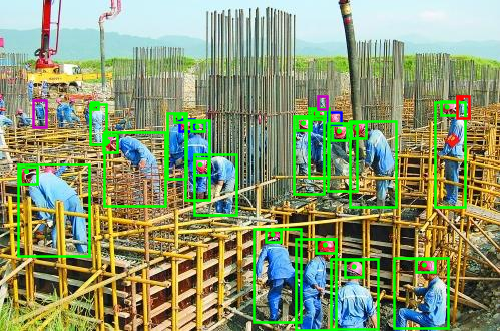}}
	    \hfill
	    \subfloat[failure in keypoint detection for partially visible person ]{\includegraphics[width=.3\textwidth]{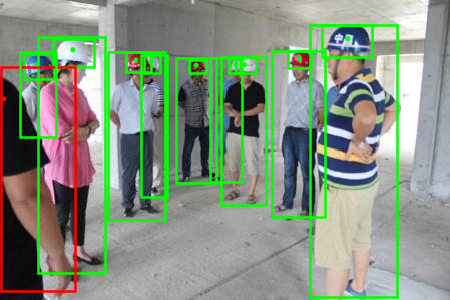}}
	\caption{Results of inference on test part of the dataset~\citep{DVN/7CBGOS_2019} achieved by best performing model (\emph{X101}), hard hat wearers with their head protection marked in green, non-wearers in red, people without head keypoint in magenta and not worn hard hats in blue.}
	\label{fig:detection_images}
\end{figure*}

The solution proposed in this paper addresses main issues observed in the hard hat detection approaches currently presented in the literature. It is based on the separate detection of people and hard hats, which means that it does not suffer from the intraclass similarity problem found in hard hat wearer/non--wearer detection. Moreover, this allows direct transfer learning from well-trained person detection models, making it easier to train and deploy. The addition of head keypoint enables the hard hat wearing to be determined with a simple, human-interpretable rule based algorithm. Moreover, no distance threshold \citep{Chen2020,Guo2020} or additional features like shoulders, hips \citep{Chen2020} or face \citep{Shen2021} are needed to establish a worker -- hard hat relationship. This alone makes our solution more flexible as it will work in more situations, where additional information will not be available for others.

In tests, it surpassed the previous solution based on the relative bounding box position of people and hard hats, as well as direct detection of hard hat wearers and non-wearers. Resulting in MS COCO style overall AP of 67.5\% compared to 66.4\% and 66.3\% achieved by the above-mentioned approaches, with class specific AP for hard hat non-wearers of 64.1\% compared to 63.0\% and 60.3\%. The performance gain in the latter task should be highlighted, as detecting workers that do not comply with the rules is the real problem.

Additionally, to fully understand performance and limitations of our solution raw detection results were also examined, some examples of images with marked object instances are presented in the Fig.~\ref{fig:detection_images}.

Most observed errors came from hard hat or person detection failures (Fig.~\ref{fig:detection_images} b, c and e). Another problem found in the results was related to the head keypoint detection. It turned out that solution detect them even for instances where head is not visible (Fig.~\ref{fig:detection_images}~f), which is caused by a very small number of such cases in dataset. However, the main surprise is the fact that head keypoints can be localized correctly even at small scale (Fig.~\ref{fig:detection_images} c and e). Causing further errors as for small-scale worker instances, matching hard hats are even smaller, making them impossible to detect.  This hinders hard hat wearing detection at small scale and whilst it would explain the worse AP\textsubscript{S} achieved by our approach in Section~\ref{sec:comparision2}. It cannot fully explain the same in Section~\ref{sec:comparision1}, as the DTs also suffer for the same exact reason, and it seems that the difference in performance is also influences by keypoint localization errors at small scale.

Moreover, it has to be mentioned that 
due to Algorithm~\ref{alg:wearing_detection} simplicity, it cannot handle significant scale difference between people and hard hats leading to a situation where a small-scale worker instance can be classified as a hard hat wearer with a hard hat at much larger scale, worn by someone else. However, this was not observed in the results.

Apart from the above, some problems with currently available datasets also have to be acknowledged. In general these datasets are significantly imbalanced, in case of the dataset used in this study the ratio between people not wearing and wearing hard hats is close to 7:3. This is clearly visible in results presented in Sections~\ref{sec:experiments} and~\ref{sec:comparision}. However, more worrying is the fact that most of the hard hat non-wearers are not workers. In the majority of cases, people instances in this group are wearing civilian clothes and the cases in which they are workers performing some tasks are rare. This should be avoided, in particular in the case of direct detection of hard hat wearers and non-wearers, as it may lead to a situation where model will learn to distinguish workers from civilians instead.

Finally, there is no benchmark dataset with multiple different labels available. Researchers tend to develop and test their solutions on custom datasets, making fair comparison impossible. Moreover, it is not uncommon to not release dataset, release partial dataset or release it without annotations. Cases in which model, model weights and means to train it are made available, are even more scarce. This is a very dangerous situation in which results presented in an article are not even replicable. To lead by an example, our solution is publicly available at \href{https://github.com/barwojcik/hard_hats}{barwojcik/hard\_hats}.

\section{Conclusions and future work}\label{sec:conclusion}

In this article, a novel approach to hard hat wearing detection based on detection of people and hard hats combined with person head keypoint localization was proposed. This unique combination provides a way to determine the correct relationship between these instances and enables differentiation between hard hat wearers and non-wearers. Results show that it surpassed both, the solution based on the relative bounding box position of people and hard hats and direct detection of hard hat wearers and non-wearers. What is even more important, main gains come from detection of hard hat non-wearers. This aspect that matters the most, as this kind of solutions should focus on detecting safety breaches. Additionally, in-depth comparisons proved that our approach does not suffer from the problem of intraclass similarity. Moreover, the addition of the person head keypoint enables the solution to be reduced to simple human-interpretable rules. Rather than an overly complex decision tree, that is unable to provide such results.

However, reliable detection of hard hat non-wearers is only the first step for development of deep learning supported safety system for construction site monitoring purposes. For such a system to be effective, workers who break the rules have to be identified, so they can be reprimanded, fined or sent to additional OHS training. Hard hat wearing detection based on face detection, like the one described by \cite{Shen2021}, is not an answer to that problem as it will not work in situations when workers face is not visible. Instead, a solution similar to one described by \cite{Zhao2019} should be considered. Face detection and identification should be done simultaneously to safety rule checking as each worker is tracked, ideally in multiple views at once. Moreover, all these tasks should be done in real-time as the construction site is a dynamic environment.

This brings us to a hardware problem, as little attention is paid to the infrastructure needed on the construction site to deploy these models in real-time. This is a very important aspect, especially considering the recent surge in GPUs. Simply moving processing to the cloud services will not be enough as real-time streaming of multiple high quality video feeds still need lightning-fast internet connection, while providing awful amount of data to analyze.

An answer to this problem could be usage of solutions aiming at efficient computation on edge devices. Recently, some methods delivering lightened deep learning architectures were proposed \citep{Blalock2020}, along with ones specifically tailored for embedded applications \citep{Howard2017,Sandler2018}. These are slowly used in recent studies, a good example of application of the latter in construction safety context is \citep{Lu2020}. The use of such algorithms together with appropriate devices would allow the creation of a distributed computing system in which each node, starting from the input, would analyze the data gradually. Therefore, decreasing data throughput needed and lowering hardware demand.

\section*{Acknowledgment}
Work presented in this study was supported by the \textit{European Union} through the \textit{European Social Fund} as a part of a \textit{Silesian University of Technology as a Centre of Modern Education based on research and innovation} project, number of grant agreement: \textbf{POWR.03.05.00 00.z098/17-00} (B.W. and M.Ż.) and as a part of a \textit{AIDA (Applied Integrative Data Analysis) - interdisciplinary doctoral studies in the field of data processing and analysis} project, number of grant agreement: \textbf{POWR.03.02.00-00-I029/17-00} (K.K.).

Authors would like to thank Liangbin Xie from Northestern University, China, for collecting and making public the hard hat dataset (doi: 10.7910/DVN/7CBGOS).

{\small
\bibliographystyle{cas-model2-names}
\bibliography{refs}

\begin{thebibliography}{52}
\expandafter\ifx\csname natexlab\endcsname\relax\def\natexlab#1{#1}\fi
\providecommand{\url}[1]{\texttt{#1}}
\providecommand{\href}[2]{#2}
\providecommand{\path}[1]{#1}
\providecommand{\DOIprefix}{doi:}
\providecommand{\ArXivprefix}{arXiv:}
\providecommand{\URLprefix}{URL: }
\providecommand{\Pubmedprefix}{pmid:}
\providecommand{\doi}[1]{\href{http://dx.doi.org/#1}{\path{#1}}}
\providecommand{\Pubmed}[1]{\href{pmid:#1}{\path{#1}}}
\providecommand{\bibinfo}[2]{#2}
\ifx\xfnm\relax \def\xfnm[#1]{\unskip,\space#1}\fi
\bibitem[{Blalock et~al.(2020)Blalock, Ortiz, Frankle and Guttag}]{Blalock2020}
\bibinfo{author}{Blalock, D.}, \bibinfo{author}{Ortiz, J.J.G.},
  \bibinfo{author}{Frankle, J.}, \bibinfo{author}{Guttag, J.},
  \bibinfo{year}{2020}.
\newblock \bibinfo{title}{What is the state of neural network pruning?}
\newblock \href{http://arxiv.org/abs/2003.03033}{\tt arXiv:2003.03033}.
\bibitem[{Brolin et~al.(2021)Brolin, Lanner and Halldin}]{BROLIN2021}
\bibinfo{author}{Brolin, K.}, \bibinfo{author}{Lanner, D.},
  \bibinfo{author}{Halldin, P.}, \bibinfo{year}{2021}.
\newblock \bibinfo{title}{Work-related traumatic brain injury in the
  construction industry in sweden and germany}.
\newblock \bibinfo{journal}{Safety Science} \bibinfo{volume}{136},
  \bibinfo{pages}{105147}.
\newblock \DOIprefix\doi{10.1016/j.ssci.2020.105147}.
\bibitem[{Cai et~al.(2017)Cai, Zuo and Zhang}]{Cai2017}
\bibinfo{author}{Cai, S.}, \bibinfo{author}{Zuo, W.}, \bibinfo{author}{Zhang,
  L.}, \bibinfo{year}{2017}.
\newblock \bibinfo{title}{Higher-order integration of hierarchical
  convolutional activations for fine-grained visual categorization}, in:
  \bibinfo{booktitle}{Proc. IEEE International Conference on Computer Vision
  (ICCV)}, pp. \bibinfo{pages}{511--520}.
\newblock \DOIprefix\doi{10.1109/ICCV.2017.63}.
\bibitem[{Chen and Demachi(2020)}]{Chen2020}
\bibinfo{author}{Chen, S.}, \bibinfo{author}{Demachi, K.},
  \bibinfo{year}{2020}.
\newblock \bibinfo{title}{A vision-based approach for ensuring proper use of
  personal protective equipment (ppe) in decommissioning of fukushima {Daiichi}
  nuclear power station}.
\newblock \bibinfo{journal}{Applied Sciences} \bibinfo{volume}{10},
  \bibinfo{pages}{5129}.
\newblock \DOIprefix\doi{10.3390/app10155129}.
\bibitem[{Colantonio et~al.(2009)Colantonio, McVittie, Lewko and
  Yin}]{Colantonio2009}
\bibinfo{author}{Colantonio, A.}, \bibinfo{author}{McVittie, D.},
  \bibinfo{author}{Lewko, J.}, \bibinfo{author}{Yin, J.}, \bibinfo{year}{2009}.
\newblock \bibinfo{title}{Traumatic brain injuries in the construction
  industry}.
\newblock \bibinfo{journal}{Brain Injury} \bibinfo{volume}{23},
  \bibinfo{pages}{873--878}.
\newblock \DOIprefix\doi{10.1080/02699050903036033}. \bibinfo{note}{pMID:
  20100123}.
\bibitem[{Colantonio et~al.(2010)Colantonio, Mroczek, Patel, Lewko, Fergenbaum
  and Brison}]{Colantonio2010}
\bibinfo{author}{Colantonio, A.}, \bibinfo{author}{Mroczek, D.},
  \bibinfo{author}{Patel, J.}, \bibinfo{author}{Lewko, J.},
  \bibinfo{author}{Fergenbaum, J.}, \bibinfo{author}{Brison, R.},
  \bibinfo{year}{2010}.
\newblock \bibinfo{title}{{Examining Occupational Traumatic Brain Injury in
  Ontario}}.
\newblock \bibinfo{journal}{Canadian Journal of Public Health}
  \bibinfo{volume}{101}, \bibinfo{pages}{S58--S62}.
\newblock \DOIprefix\doi{10.1007/BF03403848}.
\bibitem[{EU-OSHA(1989)}]{EU-OSHA1989}
\bibinfo{author}{EU-OSHA}, \bibinfo{year}{1989}.
\newblock \bibinfo{title}{{Directive 89/656/EEC - use of personal protective
  equipment}}.
\newblock \URLprefix \url{https://osha.europa.eu/en/legislation/directives/4}.
\bibitem[{Eurostat(2020)}]{Eurostat2020}
\bibinfo{author}{Eurostat}, \bibinfo{year}{2020}.
\newblock \bibinfo{title}{{Accidents at work - statistics by economic
  activity}}.
\newblock \URLprefix
  \url{https://ec.europa.eu/eurostat/statistics-explained/index.php?title=Accidents_at_work_-_statistics_by_economic_activity}.
\bibitem[{Fang et~al.(2018a)Fang, Li, Luo, Ding, Luo, Rose and An}]{Fang2018_2}
\bibinfo{author}{Fang, Q.}, \bibinfo{author}{Li, H.}, \bibinfo{author}{Luo,
  X.}, \bibinfo{author}{Ding, L.}, \bibinfo{author}{Luo, H.},
  \bibinfo{author}{Rose, T.M.}, \bibinfo{author}{An, W.},
  \bibinfo{year}{2018}a.
\newblock \bibinfo{title}{Detecting non-hardhat-use by a deep learning method
  from far-field surveillance videos}.
\newblock \bibinfo{journal}{Automation in Construction} \bibinfo{volume}{85},
  \bibinfo{pages}{1--9}.
\newblock \DOIprefix\doi{10.1016/j.autcon.2017.09.018}.
\bibitem[{Fang et~al.(2018b)Fang, Li, Luo, Ding, Rose, An and Yu}]{Fang2018_3}
\bibinfo{author}{Fang, Q.}, \bibinfo{author}{Li, H.}, \bibinfo{author}{Luo,
  X.}, \bibinfo{author}{Ding, L.}, \bibinfo{author}{Rose, T.M.},
  \bibinfo{author}{An, W.}, \bibinfo{author}{Yu, Y.}, \bibinfo{year}{2018}b.
\newblock \bibinfo{title}{A deep learning-based method for detecting
  non-certified work on construction sites}.
\newblock \bibinfo{journal}{Advanced Engineering Informatics}
  \bibinfo{volume}{35}, \bibinfo{pages}{56--68}.
\newblock \DOIprefix\doi{10.1016/j.aei.2018.01.001}.
\bibitem[{Fang et~al.(2018c)Fang, Ding, Luo and Love}]{Fang2018}
\bibinfo{author}{Fang, W.}, \bibinfo{author}{Ding, L.}, \bibinfo{author}{Luo,
  H.}, \bibinfo{author}{Love, P.E.}, \bibinfo{year}{2018}c.
\newblock \bibinfo{title}{Falls from heights: A computer vision-based approach
  for safety harness detection}.
\newblock \bibinfo{journal}{Automation in Construction} \bibinfo{volume}{91},
  \bibinfo{pages}{53--61}.
\newblock \DOIprefix\doi{10.1016/j.autcon.2018.02.018}.
\bibitem[{Fang et~al.(2020)Fang, Love, Luo and Ding}]{Fang2020}
\bibinfo{author}{Fang, W.}, \bibinfo{author}{Love, P.E.}, \bibinfo{author}{Luo,
  H.}, \bibinfo{author}{Ding, L.}, \bibinfo{year}{2020}.
\newblock \bibinfo{title}{Computer vision for behaviour-based safety in
  construction: A review and future directions}.
\newblock \bibinfo{journal}{Advanced Engineering Informatics}
  \bibinfo{volume}{43}, \bibinfo{pages}{100980}.
\newblock \DOIprefix\doi{10.1016/j.aei.2019.100980}.
\bibitem[{Fang et~al.(2019)Fang, Zhong, Zhao, Love, Luo, Xue and Xu}]{Fang2019}
\bibinfo{author}{Fang, W.}, \bibinfo{author}{Zhong, B.}, \bibinfo{author}{Zhao,
  N.}, \bibinfo{author}{Love, P.E.}, \bibinfo{author}{Luo, H.},
  \bibinfo{author}{Xue, J.}, \bibinfo{author}{Xu, S.}, \bibinfo{year}{2019}.
\newblock \bibinfo{title}{A deep learning-based approach for mitigating falls
  from height with computer vision: Convolutional neural network}.
\newblock \bibinfo{journal}{Advanced Engineering Informatics}
  \bibinfo{volume}{39}, \bibinfo{pages}{170--177}.
\newblock \DOIprefix\doi{10.1016/j.aei.2018.12.005}.
\bibitem[{Guo et~al.(2020)Guo, Li, Wang and Zhou}]{Guo2020}
\bibinfo{author}{Guo, S.}, \bibinfo{author}{Li, D.}, \bibinfo{author}{Wang,
  Z.}, \bibinfo{author}{Zhou, X.}, \bibinfo{year}{2020}.
\newblock \bibinfo{title}{Safety helmet detection method based on {Faster
  R-CNN}}, in: \bibinfo{editor}{Sun, X.}, \bibinfo{editor}{Wang, J.},
  \bibinfo{editor}{Bertino, E.} (Eds.), \bibinfo{booktitle}{ICAIS 2020:
  Artificial Intelligence and Security}, pp. \bibinfo{pages}{423--434}.
\newblock \DOIprefix\doi{10.1007/978-981-15-8086-4_40}.
\bibitem[{{Han} et~al.(2019){Han}, {Yao}, {Cheng}, {Feng} and {Xu}}]{Han2019}
\bibinfo{author}{{Han}, J.}, \bibinfo{author}{{Yao}, X.},
  \bibinfo{author}{{Cheng}, G.}, \bibinfo{author}{{Feng}, X.},
  \bibinfo{author}{{Xu}, D.}, \bibinfo{year}{2019}.
\newblock \bibinfo{title}{{P-CNN}: Part-based convolutional neural networks for
  fine-grained visual categorization}.
\newblock \bibinfo{journal}{IEEE Transactions on Pattern Analysis and Machine
  Intelligence} , \bibinfo{pages}{1}\DOIprefix\doi{10.1109/TPAMI.2019.2933510}.
\bibitem[{{He} et~al.(2017){He}, {Gkioxari}, {Dollár} and {Girshick}}]{He2017}
\bibinfo{author}{{He}, K.}, \bibinfo{author}{{Gkioxari}, G.},
  \bibinfo{author}{{Dollár}, P.}, \bibinfo{author}{{Girshick}, R.},
  \bibinfo{year}{2017}.
\newblock \bibinfo{title}{Mask r-cnn}, in: \bibinfo{booktitle}{2017 IEEE
  International Conference on Computer Vision (ICCV)}, pp.
  \bibinfo{pages}{2980--2988}.
\newblock \DOIprefix\doi{10.1109/ICCV.2017.322}.
\bibitem[{{He} et~al.(2016){He}, {Zhang}, {Ren} and {Sun}}]{He2016}
\bibinfo{author}{{He}, K.}, \bibinfo{author}{{Zhang}, X.},
  \bibinfo{author}{{Ren}, S.}, \bibinfo{author}{{Sun}, J.},
  \bibinfo{year}{2016}.
\newblock \bibinfo{title}{Deep residual learning for image recognition}, in:
  \bibinfo{booktitle}{2016 IEEE Conference on Computer Vision and Pattern
  Recognition (CVPR)}, pp. \bibinfo{pages}{770--778}.
\newblock \DOIprefix\doi{10.1109/CVPR.2016.90}.
\bibitem[{Howard et~al.(2017)Howard, Zhu, Chen, Kalenichenko, Wang, Weyand,
  Andreetto and Adam}]{Howard2017}
\bibinfo{author}{Howard, A.G.}, \bibinfo{author}{Zhu, M.},
  \bibinfo{author}{Chen, B.}, \bibinfo{author}{Kalenichenko, D.},
  \bibinfo{author}{Wang, W.}, \bibinfo{author}{Weyand, T.},
  \bibinfo{author}{Andreetto, M.}, \bibinfo{author}{Adam, H.},
  \bibinfo{year}{2017}.
\newblock \bibinfo{title}{{MobileNets}: Efficient convolutional neural networks
  for mobile vision applications}.
\newblock \href{http://arxiv.org/abs/1704.04861}{\tt arXiv:1704.04861}.
\bibitem[{Kelm et~al.(2013)Kelm, Laußat, Meins-Becker, Platz, Khazaee, Costin,
  Helmus and Teizer}]{Kelm2013}
\bibinfo{author}{Kelm, A.}, \bibinfo{author}{Laußat, L.},
  \bibinfo{author}{Meins-Becker, A.}, \bibinfo{author}{Platz, D.},
  \bibinfo{author}{Khazaee, M.}, \bibinfo{author}{Costin, A.},
  \bibinfo{author}{Helmus, M.}, \bibinfo{author}{Teizer, J.},
  \bibinfo{year}{2013}.
\newblock \bibinfo{title}{Mobile passive radio frequency identification
  {(RFID)} portal for automated and rapid control of personal protective
  equipment {(PPE)} on construction sites}.
\newblock \bibinfo{journal}{Automation in Construction} \bibinfo{volume}{36},
  \bibinfo{pages}{38–52}.
\newblock \DOIprefix\doi{10.1016/j.autcon.2013.08.009}.
\bibitem[{Khan et~al.(2021)Khan, Saleem, Lee, Park and Park}]{KHAN2021}
\bibinfo{author}{Khan, N.}, \bibinfo{author}{Saleem, M.R.},
  \bibinfo{author}{Lee, D.}, \bibinfo{author}{Park, M.W.},
  \bibinfo{author}{Park, C.}, \bibinfo{year}{2021}.
\newblock \bibinfo{title}{Utilizing safety rule correlation for mobile
  scaffolds monitoring leveraging deep convolution neural networks}.
\newblock \bibinfo{journal}{Computers in Industry} \bibinfo{volume}{129},
  \bibinfo{pages}{103448}.
\newblock \DOIprefix\doi{10.1016/j.compind.2021.103448}.
\bibitem[{Konda et~al.(2016)Konda, Tiesman and Reichard}]{Konda2016}
\bibinfo{author}{Konda, S.}, \bibinfo{author}{Tiesman, H.M.},
  \bibinfo{author}{Reichard, A.A.}, \bibinfo{year}{2016}.
\newblock \bibinfo{title}{Fatal traumatic brain injuries in the construction
  industry, 2003-2010}.
\newblock \bibinfo{journal}{American Journal of Industrial Medicine}
  \bibinfo{volume}{59}, \bibinfo{pages}{212--220}.
\newblock \DOIprefix\doi{10.1002/ajim.22557}.
\bibitem[{{Li} et~al.(2019){Li}, {Wang}, {Wang}, {Tai}, {Qian}, {Yang}, {Wang},
  {Li} and {Huang}}]{Li2019}
\bibinfo{author}{{Li}, J.}, \bibinfo{author}{{Wang}, Y.},
  \bibinfo{author}{{Wang}, C.}, \bibinfo{author}{{Tai}, Y.},
  \bibinfo{author}{{Qian}, J.}, \bibinfo{author}{{Yang}, J.},
  \bibinfo{author}{{Wang}, C.}, \bibinfo{author}{{Li}, J.},
  \bibinfo{author}{{Huang}, F.}, \bibinfo{year}{2019}.
\newblock \bibinfo{title}{{DSFD:} dual shot face detector}, in:
  \bibinfo{booktitle}{2019 IEEE/CVF Conference on Computer Vision and Pattern
  Recognition (CVPR)}, pp. \bibinfo{pages}{5055--5064}.
\newblock \DOIprefix\doi{10.1109/CVPR.2019.00520}.
\bibitem[{{Lin} et~al.(2017){Lin}, {Dollár}, {Girshick}, {He}, {Hariharan} and
  {Belongie}}]{Lin2017}
\bibinfo{author}{{Lin}, T.}, \bibinfo{author}{{Dollár}, P.},
  \bibinfo{author}{{Girshick}, R.}, \bibinfo{author}{{He}, K.},
  \bibinfo{author}{{Hariharan}, B.}, \bibinfo{author}{{Belongie}, S.},
  \bibinfo{year}{2017}.
\newblock \bibinfo{title}{Feature pyramid networks for object detection}, in:
  \bibinfo{booktitle}{2017 IEEE Conference on Computer Vision and Pattern
  Recognition (CVPR)}, pp. \bibinfo{pages}{936--944}.
\newblock \DOIprefix\doi{10.1109/CVPR.2017.106}.
\bibitem[{Lin et~al.(2015)Lin, Maire, Belongie, Bourdev, Girshick, Hays,
  Perona, Ramanan, Zitnick and Dollár}]{coco2015}
\bibinfo{author}{Lin, T.Y.}, \bibinfo{author}{Maire, M.},
  \bibinfo{author}{Belongie, S.}, \bibinfo{author}{Bourdev, L.},
  \bibinfo{author}{Girshick, R.}, \bibinfo{author}{Hays, J.},
  \bibinfo{author}{Perona, P.}, \bibinfo{author}{Ramanan, D.},
  \bibinfo{author}{Zitnick, C.L.}, \bibinfo{author}{Dollár, P.},
  \bibinfo{year}{2015}.
\newblock \bibinfo{title}{Microsoft {COCO}: {Common Objects in Context}}.
\newblock \href{http://arxiv.org/abs/1405.0312}{\tt arXiv:1405.0312}.
\bibitem[{Liu et~al.(2011)Liu, Wei, Salerno, Comper and Colantonio}]{Liu2011}
\bibinfo{author}{Liu, M.}, \bibinfo{author}{Wei, W.}, \bibinfo{author}{Salerno,
  J.}, \bibinfo{author}{Comper, P.}, \bibinfo{author}{Colantonio, A.},
  \bibinfo{year}{2011}.
\newblock \bibinfo{title}{Work-related mild-moderate traumatic brain injury and
  the construction industry}.
\newblock \bibinfo{journal}{Work (Reading, Mass.)} \bibinfo{volume}{39},
  \bibinfo{pages}{283--90}.
\newblock \DOIprefix\doi{10.3233/WOR-2011-1176}.
\bibitem[{Liu et~al.(2016)Liu, Anguelov, Erhan, Szegedy, Reed, Fu and
  Berg}]{Liu2016}
\bibinfo{author}{Liu, W.}, \bibinfo{author}{Anguelov, D.},
  \bibinfo{author}{Erhan, D.}, \bibinfo{author}{Szegedy, C.},
  \bibinfo{author}{Reed, S.}, \bibinfo{author}{Fu, C.Y.},
  \bibinfo{author}{Berg, A.C.}, \bibinfo{year}{2016}.
\newblock \bibinfo{title}{{SSD}: Single shot multibox detector}, in:
  \bibinfo{booktitle}{Computer Vision -- {ECCV 2016}}, pp.
  \bibinfo{pages}{21--37}.
\newblock \DOIprefix\doi{10.1007/978-3-319-46448-0_2}.
\bibitem[{Luo et~al.(2020)Luo, Liu, Fang, Love, Yu and Lu}]{LuoH2020}
\bibinfo{author}{Luo, H.}, \bibinfo{author}{Liu, J.}, \bibinfo{author}{Fang,
  W.}, \bibinfo{author}{Love, P.E.}, \bibinfo{author}{Yu, Q.},
  \bibinfo{author}{Lu, Z.}, \bibinfo{year}{2020}.
\newblock \bibinfo{title}{Real-time smart video surveillance to manage safety:
  A case study of a transport mega-project}.
\newblock \bibinfo{journal}{Advanced Engineering Informatics}
  \bibinfo{volume}{45}, \bibinfo{pages}{101100}.
\newblock \DOIprefix\doi{10.1016/j.aei.2020.101100}.
\bibitem[{{Luo} et~al.(2019){Luo}, {Yang}, {Mo}, {Lu}, {Davis}, {Li}, {Yang}
  and {Lim}}]{Lou2019}
\bibinfo{author}{{Luo}, W.}, \bibinfo{author}{{Yang}, X.},
  \bibinfo{author}{{Mo}, X.}, \bibinfo{author}{{Lu}, Y.},
  \bibinfo{author}{{Davis}, L.}, \bibinfo{author}{{Li}, J.},
  \bibinfo{author}{{Yang}, J.}, \bibinfo{author}{{Lim}, S.},
  \bibinfo{year}{2019}.
\newblock \bibinfo{title}{Cross-x learning for fine-grained visual
  categorization}, in: \bibinfo{booktitle}{2019 IEEE/CVF International
  Conference on Computer Vision (ICCV)}, pp. \bibinfo{pages}{8241--8250}.
\newblock \DOIprefix\doi{10.1109/ICCV.2019.00833}.
\bibitem[{Memarzadeh et~al.(2012)Memarzadeh, Heydarian, Golparvar-Fard and
  Niebles}]{Memarzadeh2012}
\bibinfo{author}{Memarzadeh, M.}, \bibinfo{author}{Heydarian, A.},
  \bibinfo{author}{Golparvar-Fard, M.}, \bibinfo{author}{Niebles, J.C.},
  \bibinfo{year}{2012}.
\newblock \bibinfo{title}{Real-time and automated 2d recognition and tracking
  of workers and equipment from site video streams for construction performance
  assessment}.
\newblock \bibinfo{journal}{Congress on Computing in Civil Engineering,
  Proceedings} \DOIprefix\doi{10.1061/9780784412343.0054}.
\bibitem[{Mneymneh et~al.(2019)Mneymneh, Abbas and Khoury}]{Mneymneh2019}
\bibinfo{author}{Mneymneh, B.E.}, \bibinfo{author}{Abbas, M.},
  \bibinfo{author}{Khoury, H.}, \bibinfo{year}{2019}.
\newblock \bibinfo{title}{Vision-based framework for intelligent monitoring of
  hardhat wearing on construction sites}.
\newblock \bibinfo{journal}{Journal of Computing in Civil Engineering}
  \bibinfo{volume}{33}, \bibinfo{pages}{04018066}.
\newblock \DOIprefix\doi{10.1061/(ASCE)CP.1943-5487.0000813}.
\bibitem[{Nath et~al.(2020)Nath, Behzadan and Paal}]{Nath2020}
\bibinfo{author}{Nath, N.D.}, \bibinfo{author}{Behzadan, A.H.},
  \bibinfo{author}{Paal, S.G.}, \bibinfo{year}{2020}.
\newblock \bibinfo{title}{Deep learning for site safety: Real-time detection of
  personal protective equipment}.
\newblock \bibinfo{journal}{Automation in Construction} \bibinfo{volume}{112},
  \bibinfo{pages}{103085}.
\newblock \DOIprefix\doi{10.1016/j.autcon.2020.103085}.
\bibitem[{Park et~al.(2015)Park, Elsafty and Zhu}]{Park2015}
\bibinfo{author}{Park, M.W.}, \bibinfo{author}{Elsafty, N.},
  \bibinfo{author}{Zhu, Z.}, \bibinfo{year}{2015}.
\newblock \bibinfo{title}{Hardhat-wearing detection for enhancing on-site
  safety of construction workers}.
\newblock \bibinfo{journal}{Journal of Construction Engineering and Management}
  \bibinfo{volume}{141}, \bibinfo{pages}{04015024}.
\newblock \DOIprefix\doi{10.1061/(ASCE)CO.1943-7862.0000974}.
\bibitem[{Redmon and Farhadi(2018)}]{Redmon&Farhadi2018}
\bibinfo{author}{Redmon, J.}, \bibinfo{author}{Farhadi, A.},
  \bibinfo{year}{2018}.
\newblock \bibinfo{title}{Yolov3: An incremental improvement}.
\newblock \href{http://arxiv.org/abs/1804.02767}{\tt arXiv:1804.02767}.
\bibitem[{{Ren} et~al.(2017){Ren}, {He}, {Girshick} and {Sun}}]{Ren2017}
\bibinfo{author}{{Ren}, S.}, \bibinfo{author}{{He}, K.},
  \bibinfo{author}{{Girshick}, R.}, \bibinfo{author}{{Sun}, J.},
  \bibinfo{year}{2017}.
\newblock \bibinfo{title}{Faster r-cnn: Towards real-time object detection with
  region proposal networks}.
\newblock \bibinfo{journal}{IEEE Transactions on Pattern Analysis and Machine
  Intelligence} \bibinfo{volume}{39}, \bibinfo{pages}{1137--1149}.
\newblock \DOIprefix\doi{10.1109/TPAMI.2016.2577031}.
\bibitem[{Salem et~al.(2013)Salem, Jaumally, Bayanzay, Khoury and
  Torkaman}]{Salem2013}
\bibinfo{author}{Salem, A.M.O.}, \bibinfo{author}{Jaumally, B.A.},
  \bibinfo{author}{Bayanzay, K.}, \bibinfo{author}{Khoury, K.},
  \bibinfo{author}{Torkaman, A.}, \bibinfo{year}{2013}.
\newblock \bibinfo{title}{{Traumatic brain injuries from work accidents: a
  retrospective study}}.
\newblock \bibinfo{journal}{Occupational Medicine} \bibinfo{volume}{63},
  \bibinfo{pages}{358--360}.
\newblock \DOIprefix\doi{10.1093/occmed/kqt037}.
\bibitem[{Sandler et~al.(2018)Sandler, Howard, Zhu, Zhmoginov and
  Chen}]{Sandler2018}
\bibinfo{author}{Sandler, M.}, \bibinfo{author}{Howard, A.},
  \bibinfo{author}{Zhu, M.}, \bibinfo{author}{Zhmoginov, A.},
  \bibinfo{author}{Chen, L.C.}, \bibinfo{year}{2018}.
\newblock \bibinfo{title}{{MobileNetV2}: Inverted residuals and linear
  bottlenecks}, in: \bibinfo{booktitle}{Proceedings of the IEEE Conference on
  Computer Vision and Pattern Recognition (CVPR)}, pp.
  \bibinfo{pages}{4510--4520}.
\newblock \DOIprefix\doi{10.1109/CVPR.2018.00474}.
\bibitem[{Sharma et~al.(2019)Sharma, Nowrouzi-Kia, Mollayeva, Kontos,
  Grigorovich, Liss, Gibson, Mantis, Lewko and Colantonio}]{Bhanu2019}
\bibinfo{author}{Sharma, B.}, \bibinfo{author}{Nowrouzi-Kia, B.},
  \bibinfo{author}{Mollayeva, T.}, \bibinfo{author}{Kontos, P.},
  \bibinfo{author}{Grigorovich, A.}, \bibinfo{author}{Liss, G.},
  \bibinfo{author}{Gibson, B.}, \bibinfo{author}{Mantis, S.},
  \bibinfo{author}{Lewko, J.}, \bibinfo{author}{Colantonio, A.},
  \bibinfo{year}{2019}.
\newblock \bibinfo{title}{Work-related traumatic brain injury: A brief report
  on workers perspective on job and health and safety training, supervision,
  and injury preventability}.
\newblock \bibinfo{journal}{Work} \bibinfo{volume}{62},
  \bibinfo{pages}{319--325}.
\newblock \DOIprefix\doi{10.3233/WOR-192866}.
\bibitem[{Shen et~al.(2021)Shen, Xiong, Li, He, Li and Zheng}]{Shen2021}
\bibinfo{author}{Shen, J.}, \bibinfo{author}{Xiong, X.}, \bibinfo{author}{Li,
  Y.}, \bibinfo{author}{He, W.}, \bibinfo{author}{Li, P.},
  \bibinfo{author}{Zheng, X.}, \bibinfo{year}{2021}.
\newblock \bibinfo{title}{Detecting safety helmet wearing on construction sites
  with bounding-box regression and deep transfer learning}.
\newblock \bibinfo{journal}{Computer-Aided Civil and Infrastructure
  Engineering} \bibinfo{volume}{36}, \bibinfo{pages}{180--196}.
\newblock \DOIprefix\doi{10.1111/mice.12579}.
\bibitem[{Tang et~al.(2020)Tang, Roberts and Golparvar-Fard}]{Tang2020}
\bibinfo{author}{Tang, S.}, \bibinfo{author}{Roberts, D.},
  \bibinfo{author}{Golparvar-Fard, M.}, \bibinfo{year}{2020}.
\newblock \bibinfo{title}{Human-object interaction recognition for automatic
  construction site safety inspection}.
\newblock \bibinfo{journal}{Automation in Construction} \bibinfo{volume}{120},
  \bibinfo{pages}{103356}.
\newblock \DOIprefix\doi{10.1016/j.autcon.2020.103356}.
\bibitem[{Taylor et~al.(2017)Taylor, Bell, Breiding and Xu}]{Taylor2017}
\bibinfo{author}{Taylor, C.A.}, \bibinfo{author}{Bell, J.M.},
  \bibinfo{author}{Breiding, M.J.}, \bibinfo{author}{Xu, L.},
  \bibinfo{year}{2017}.
\newblock \bibinfo{title}{Traumatic brain injury{\textendash}related emergency
  department visits, hospitalizations, and deaths {\textemdash} united states,
  2007 and 2013}.
\newblock \bibinfo{journal}{{MMWR}. Surveillance Summaries}
  \bibinfo{volume}{66}, \bibinfo{pages}{1--16}.
\newblock \DOIprefix\doi{10.15585/mmwr.ss6609a1}.
\bibitem[{Tun et~al.(2020)Tun, Kim, Jeon, Kim and Lee}]{WaiKim2020}
\bibinfo{author}{Tun, W.}, \bibinfo{author}{Kim, J.H.}, \bibinfo{author}{Jeon,
  Y.}, \bibinfo{author}{Kim, S.}, \bibinfo{author}{Lee, J.W.},
  \bibinfo{year}{2020}.
\newblock \bibinfo{title}{Safety helmet and vest wearing detection approach by
  integrating yolo and svm for uav}, in: \bibinfo{booktitle}{Korean Society for
  Aeronautical and Space Sciences 2020 Spring Conference}.
\newblock \URLprefix
  \url{https://www.dbpia.co.kr/Journal/articleDetail?nodeId=NODE10442178}.
\bibitem[{{United States Bureau of Labor Statistics}(2020)}]{BLS2020}
\bibinfo{author}{{United States Bureau of Labor Statistics}},
  \bibinfo{year}{2020}.
\newblock \bibinfo{title}{{National census of fatal occupational injuries in
  2019}}.
\newblock \URLprefix \url{https://www.bls.gov/news.release/cfoi.nr0.htm}.
\bibitem[{{United States Department of Labor}(2004)}]{OSHA2004}
\bibinfo{author}{{United States Department of Labor}}, \bibinfo{year}{2004}.
\newblock \bibinfo{title}{{Personal Protective Equipment}}.
\bibitem[{Wang et~al.(2020)Wang, Xie, Yang, Deng, Du and Xu}]{Lu2020}
\bibinfo{author}{Wang, L.}, \bibinfo{author}{Xie, L.}, \bibinfo{author}{Yang,
  P.}, \bibinfo{author}{Deng, Q.}, \bibinfo{author}{Du, S.},
  \bibinfo{author}{Xu, L.}, \bibinfo{year}{2020}.
\newblock \bibinfo{title}{Hardhat-wearing detection based on a lightweight
  convolutional neural network with multi-scale features and a top-down
  module}.
\newblock \bibinfo{journal}{Sensors} \bibinfo{volume}{20}.
\newblock \DOIprefix\doi{10.3390/s20071868}.
\bibitem[{Wei et~al.(2019)Wei, Love, Fang, Luo and Xu}]{Wei2019}
\bibinfo{author}{Wei, R.}, \bibinfo{author}{Love, P.E.}, \bibinfo{author}{Fang,
  W.}, \bibinfo{author}{Luo, H.}, \bibinfo{author}{Xu, S.},
  \bibinfo{year}{2019}.
\newblock \bibinfo{title}{Recognizing people’s identity in construction sites
  with computer vision: A spatial and temporal attention pooling network}.
\newblock \bibinfo{journal}{Advanced Engineering Informatics}
  \bibinfo{volume}{42}, \bibinfo{pages}{100981}.
\newblock \DOIprefix\doi{10.1016/j.aei.2019.100981}.
\bibitem[{Wu et~al.(2019a)Wu, Cai, Chen, Wang and Wang}]{Wu2019}
\bibinfo{author}{Wu, J.}, \bibinfo{author}{Cai, N.}, \bibinfo{author}{Chen,
  W.}, \bibinfo{author}{Wang, H.}, \bibinfo{author}{Wang, G.},
  \bibinfo{year}{2019}a.
\newblock \bibinfo{title}{Automatic detection of hardhats worn by construction
  personnel: A deep learning approach and benchmark dataset}.
\newblock \bibinfo{journal}{Automation in Construction} \bibinfo{volume}{106},
  \bibinfo{pages}{102894}.
\newblock \DOIprefix\doi{10.1016/j.autcon.2019.102894}.
\bibitem[{Wu et~al.(2017)Wu, Zheng, Zhao, Li, Yan, Liang, Wang, Zhou, Lin, Fu,
  Wang and Wang}]{wu2017ai}
\bibinfo{author}{Wu, J.}, \bibinfo{author}{Zheng, H.}, \bibinfo{author}{Zhao,
  B.}, \bibinfo{author}{Li, Y.}, \bibinfo{author}{Yan, B.},
  \bibinfo{author}{Liang, R.}, \bibinfo{author}{Wang, W.},
  \bibinfo{author}{Zhou, S.}, \bibinfo{author}{Lin, G.}, \bibinfo{author}{Fu,
  Y.}, \bibinfo{author}{Wang, Y.}, \bibinfo{author}{Wang, Y.},
  \bibinfo{year}{2017}.
\newblock \bibinfo{title}{{AI Challenger}: A large-scale dataset for going
  deeper in image understanding}.
\newblock \href{http://arxiv.org/abs/1711.06475}{\tt arXiv:1711.06475}.
\bibitem[{Wu et~al.(2019b)Wu, Kirillov, Massa, Lo and
  Girshick}]{wu2019detectron2}
\bibinfo{author}{Wu, Y.}, \bibinfo{author}{Kirillov, A.},
  \bibinfo{author}{Massa, F.}, \bibinfo{author}{Lo, W.Y.},
  \bibinfo{author}{Girshick, R.}, \bibinfo{year}{2019}b.
\newblock \bibinfo{title}{Detectron2}.
\newblock \URLprefix \url{https://github.com/facebookresearch/detectron2}.
\bibitem[{Xie(2019)}]{DVN/7CBGOS_2019}
\bibinfo{author}{Xie, L.}, \bibinfo{year}{2019}.
\newblock \bibinfo{title}{{Hardhat}}.
\newblock \DOIprefix\doi{10.7910/DVN/7CBGOS}.
\bibitem[{{Xie} et~al.(2017){Xie}, {Girshick}, {Dollár}, {Tu} and
  {He}}]{Xie2017}
\bibinfo{author}{{Xie}, S.}, \bibinfo{author}{{Girshick}, R.},
  \bibinfo{author}{{Dollár}, P.}, \bibinfo{author}{{Tu}, Z.},
  \bibinfo{author}{{He}, K.}, \bibinfo{year}{2017}.
\newblock \bibinfo{title}{Aggregated residual transformations for deep neural
  networks}, in: \bibinfo{booktitle}{2017 IEEE Conference on Computer Vision
  and Pattern Recognition (CVPR)}, pp. \bibinfo{pages}{5987--5995}.
\newblock \DOIprefix\doi{10.1109/CVPR.2017.634}.
\bibitem[{{Zhao} et~al.(2019){Zhao}, {Chen}, {Cao}, {Yang}, {Xiong} and
  {Gui}}]{Zhao2019}
\bibinfo{author}{{Zhao}, Y.}, \bibinfo{author}{{Chen}, Q.},
  \bibinfo{author}{{Cao}, W.}, \bibinfo{author}{{Yang}, J.},
  \bibinfo{author}{{Xiong}, J.}, \bibinfo{author}{{Gui}, G.},
  \bibinfo{year}{2019}.
\newblock \bibinfo{title}{Deep learning for risk detection and trajectory
  tracking at construction sites}.
\newblock \bibinfo{journal}{IEEE Access} \bibinfo{volume}{7},
  \bibinfo{pages}{30905--30912}.
\newblock \DOIprefix\doi{10.1109/ACCESS.2019.2902658}.
\bibitem[{{Zhou} et~al.(2021){Zhou}, {Zhao} and {Nie}}]{Zhou2021}
\bibinfo{author}{{Zhou}, F.}, \bibinfo{author}{{Zhao}, H.},
  \bibinfo{author}{{Nie}, Z.}, \bibinfo{year}{2021}.
\newblock \bibinfo{title}{Safety helmet detection based on yolov5}, in:
  \bibinfo{booktitle}{2021 IEEE International Conference on Power Electronics,
  Computer Applications (ICPECA)}, pp. \bibinfo{pages}{6--11}.
\newblock \DOIprefix\doi{10.1109/ICPECA51329.2021.9362711}.

\end{thebibliography}
}

\end{document}